\documentclass{article}

\PassOptionsToPackage{numbers, compress}{natbib}
\usepackage[preprint]{neurips_2023}

\usepackage[utf8]{inputenc} % allow utf-8 input
\usepackage[T1]{fontenc}    % use 8-bit T1 fonts
\usepackage{hyperref}       % hyperlinks
\usepackage{url}            % simple URL typesetting
\usepackage{booktabs}       % professional-quality tables
\usepackage{amsfonts}       % blackboard math symbols
\usepackage{nicefrac}       % compact symbols for 1/2, etc.
\usepackage{microtype}      % microtypography
\usepackage{xcolor}         % colors
\usepackage{graphicx}
\usepackage{color}
\usepackage{wrapfig}
\usepackage{epsfig}
\usepackage{graphicx}
\usepackage{amsmath}
\usepackage{amssymb}
\usepackage{multirow}
\usepackage{algorithm}
\usepackage{algorithmic}
\usepackage{arydshln}
\usepackage{threeparttable}
\usepackage{colortbl}
\usepackage{enumitem}
\usepackage{siunitx}
\usepackage{bm}
\usepackage{array}
\usepackage{colortbl}
\usepackage{caption}
\usepackage{dblfloatfix}
\usepackage{amssymb}
\usepackage{pifont}
\usepackage[capitalize]{cleveref}
\usepackage{amssymb}
\usepackage{float}

\usepackage{listings}
\usepackage[export]{adjustbox}
\usepackage{makecell}
\usepackage{centernot}
\usepackage{subcaption}
\usepackage{bm}
\usepackage{bbding}
\usepackage{wrapfig}
\usepackage{capt-of}
\usepackage{enumitem}

\newcommand\figcaption{\def\@captype{figure}\caption}
\newcommand\tabcaption{\def\@captype{table}\caption}\makeatother
\definecolor{dark-gray}{gray}{0.20}
\definecolor{mygreen}{HTML}{39b54a}

\newcommand{\pub}[1]{{\color{dark-gray}{\tiny{[{#1}]}}}}

\newcolumntype{x}[1]{>{\centering\arraybackslash}p{#1pt}}
\newcolumntype{y}[1]{>{\raggedright\arraybackslash}p{#1pt}}
\newcolumntype{z}[1]{>{\raggedleft\arraybackslash}p{#1pt}}
\newcommand{\boldres}[1]{{\textbf{\textcolor{black}{#1}}}}
\newcommand{\secondres}[1]{{{\textcolor{black}{#1}}}}
\title{Meta-Transformer: A Unified Framework for Multimodal Learning}

\author{
	\textbf{Yiyuan Zhang}$^{1,2}$\thanks{Equal contribution}
	~~~ \textbf{Kaixiong Gong}$^{1,2\ast}$
	~~~ \textbf{Kaipeng Zhang}$^{2}$\thanks{Corresponding authors}\\
	~~~ \textbf{Hongsheng Li}$^{1}$
	~~~ \textbf{Yu Qiao}$^{2}$
	~~~ \textbf{Wanli Ouyang}$^{2}$
	~~~ \textbf{Xiangyu Yue}$^{1\dag}$\thanks{Project leader} \\
	\textsuperscript{1}Multimedia Lab, The Chinese University of Hong Kong~~~
	\textsuperscript{2}Shanghai AI Lab~~~ \\
	{\tt\small yiyuanzhang.ai@gmail.com, kaixionggong@gmail.com, }
	{\tt\small xyyue@ie.cuhk.edu.hk} \\
	\url{https://kxgong.github.io/meta_transformer/} \\
}

\begin{document}
\maketitle
% \let\thefootnote\relax\footnotetext{\Envelope \, denotes corresponding authors.}
%--------------Teaser Figure ----------------------------------
\vspace{-0.25in}
\begin{figure*}[ht]
	\centering
	\includegraphics[width=0.78\linewidth]{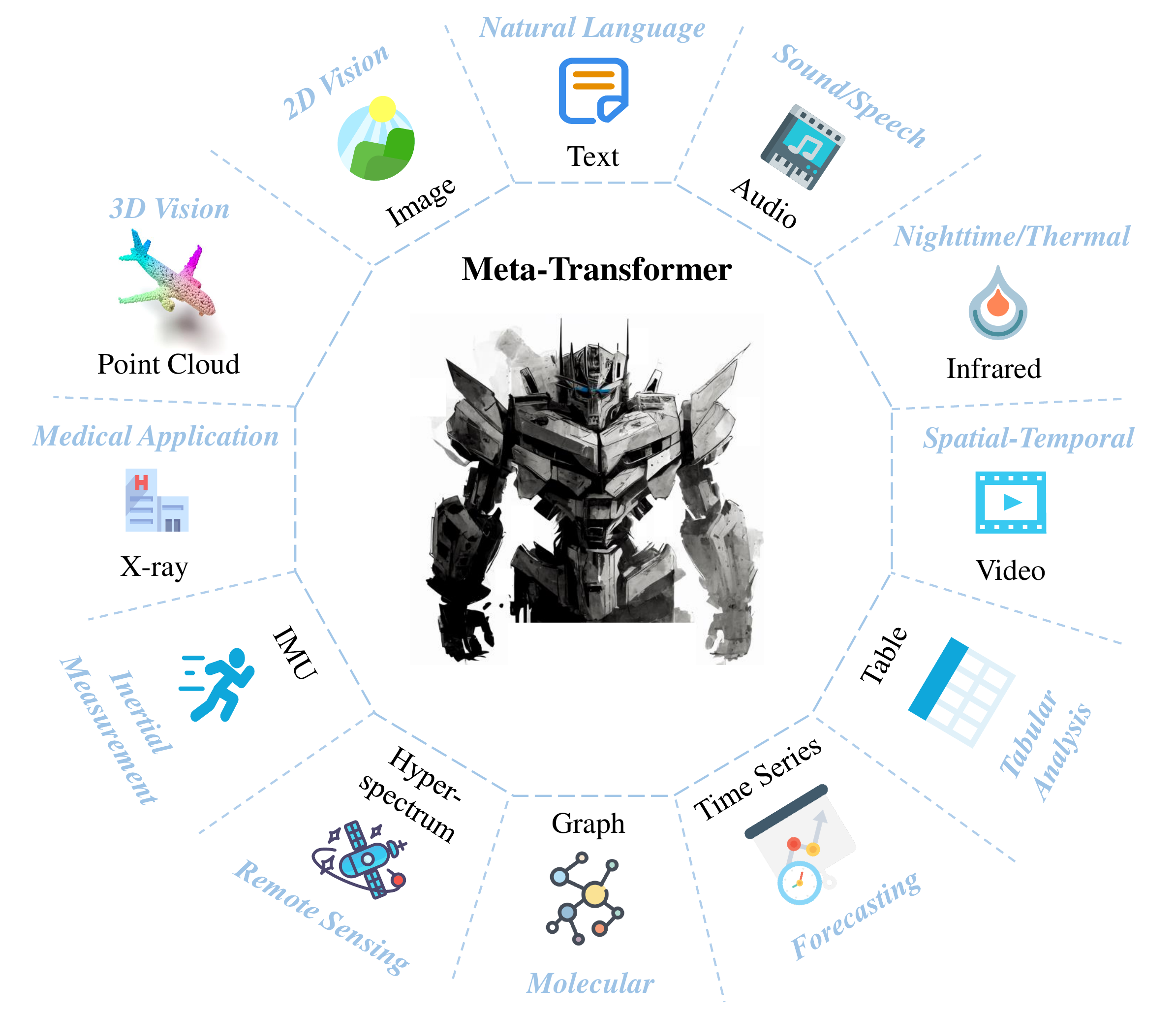}
	\caption{\textbf{Unified Multimodal Learning}. Meta-Transformer utilizes the same backbone to encode natural language, image, point cloud, audio, video, infrared, hyperspectral, X-ray, time-series, tabular, Inertial Measurement Unit (IMU), and graph data. It reveals the potential of transformer architectures for unified multi-modal intelligence.
		% universal perception.
	}
	\label{fig:teaser}
\end{figure*}
%----------------------------------------------------
\begin{abstract}
	Multimodal learning aims to build models that can process and relate information from multiple modalities. Despite years of development in this field, it still remains challenging to design a unified network for processing various modalities (\textit{e.g.} natural language, 2D images, 3D point clouds, audio, video, time series, tabular data) 
	due to the inherent gaps among them. 
	In this work, we propose a framework, named Meta-Transformer, that leverages a \textbf{frozen} encoder to perform multimodal perception without any paired multimodal training data. In Meta-Transformer, the raw input data from various modalities are mapped into a shared token space, allowing a subsequent encoder with frozen parameters to extract high-level semantic features of the input data. Composed of three main components: a unified data tokenizer, a modality-shared encoder, and task-specific heads for downstream tasks, Meta-Transformer is the first framework to perform unified learning across 12 modalities with unpaired data. 
	Experiments on different benchmarks reveal that Meta-Transformer can handle a wide range of tasks including fundamental perception (text, image, point cloud, audio, video), practical application (X-Ray, infrared, hyperspectral, and IMU), and data mining (graph, tabular, and time-series).  
	Meta-Transformer indicates a promising future for developing unified multimodal intelligence with transformers. Code will be available at \url{https://github.com/invictus717/MetaTransformer}.
\end{abstract}

% --------------------------------------------------------------------------
\section{Introduction}~\label{sec:intro}
The human brain, which is considered as the inspiration for neural network models, processes information from various sensory inputs, \textit{e.g.} visual, auditory, and tactile signals, simultaneously. Moreover, knowledge from one source can benefit the comprehension of another. However, in deep learning, designing a unified network capable of processing a wide range of data formats is a non-trivial task due to the significant modality gap~\cite{wang2021simvlm,wang2021vlmo,wang2022image}.

Each data modality presents unique data patterns, which makes it difficult to adapt models trained on one modality to another. For instance, images exhibit a high degree of information redundancy due to densely packed pixels, which is not the case with natural language~\cite{he2022masked}. Point clouds, on the other hand, have a sparse distribution in 3D space, making them more susceptible to noise and challenging to represent~\cite{qi2017pointnet++}. Audio spectrograms are time-varying and non-stationary data patterns consisting of combinations of waves across frequency domains~\cite{gong2021ast}. Video data contains a sequence of image frames, which gives it the unique capability to capture both spatial information and temporal dynamics~\cite{gberta_2021_ICML}. 
Graph data represents entities as nodes and relationships as edges in a graph, modeling complex, many-to-many relationships between entities~\cite{gilmer2017neural}. 
Owing to the substantial differences inherent to various data modalities, it is common practice to utilize distinct network architectures to encode each modality separately.
For instance, Point Transformer~\cite{zhao2021pointtransformer} leverages vector-level position attention to extract structural information from 3D coordinates, but it cannot encode an image, a natural language paragraph, or an audio spectrogram slice. 
Therefore, designing a unified framework capable of utilizing a modality-shared parameter space to encode multiple data modalities remains a significant challenge. 
Recently, the development of unified frameworks such as VLMO~\cite{wang2021vlmo}, OFA~\cite{wang2022unifying}, and BEiT-3~\cite{wang2022image} have improved the ability of the network for multimodal understanding, through large-scale multimodal pretraining on paired data~\cite{wang2022image,wang2022unifying,wang2021vlmo}, but they are more focused on vision and language, and unable to share the whole encoder across modalities

The transformer architecture and attention mechanism, proposed by Vaswani \textit{et al.} in 2017~\cite{vaswani2017attention} for natural language processing (NLP), 
have made a significant difference in deep learning~\cite{vaswani2017attention,detr,dosovitskiy2020image,zhai2021scalingvit,xie2021segformer,wang2021pvtv2}. These advancements have been instrumental in enhancing perception across different modalities such as 2D vision (including ViT~\cite{dosovitskiy2021image,chen2022vision} and Swin Transformer~\cite{liu2021swin}), 3D vision (such as Point Transformer~\cite{zhao2021pointtransformer} and Point-ViT~\cite{yu2022pointbert,qian2022pix4point}), and audio signal processing ( AST~\cite{gong2021ast}), \textit{etc}. These works have demonstrated the versatility of transformer-based architectures, inspiring researchers to explore \textit{whether it's possible to develop foundation models capable of unifying multiple modalities, ultimately achieving human-level perception across all modalities.}
\begin{table}[ht]
	\centering
	\vspace{-3mm}
	\caption{Comparison between Meta-Transformer and related works on perception tasks. 
	}
	
	\resizebox{1.0\linewidth}{!}{
		\begin{tabular}{l|c|c|c}
			\toprule
			Method & {Modalities}  & Share Parameters & Unpaired Data\\ \hline
			Transformer~\cite{vaswani2017attention} 
			& \includegraphics[height=2.8mm, trim={0cm 3cm 0cm -4cm}]{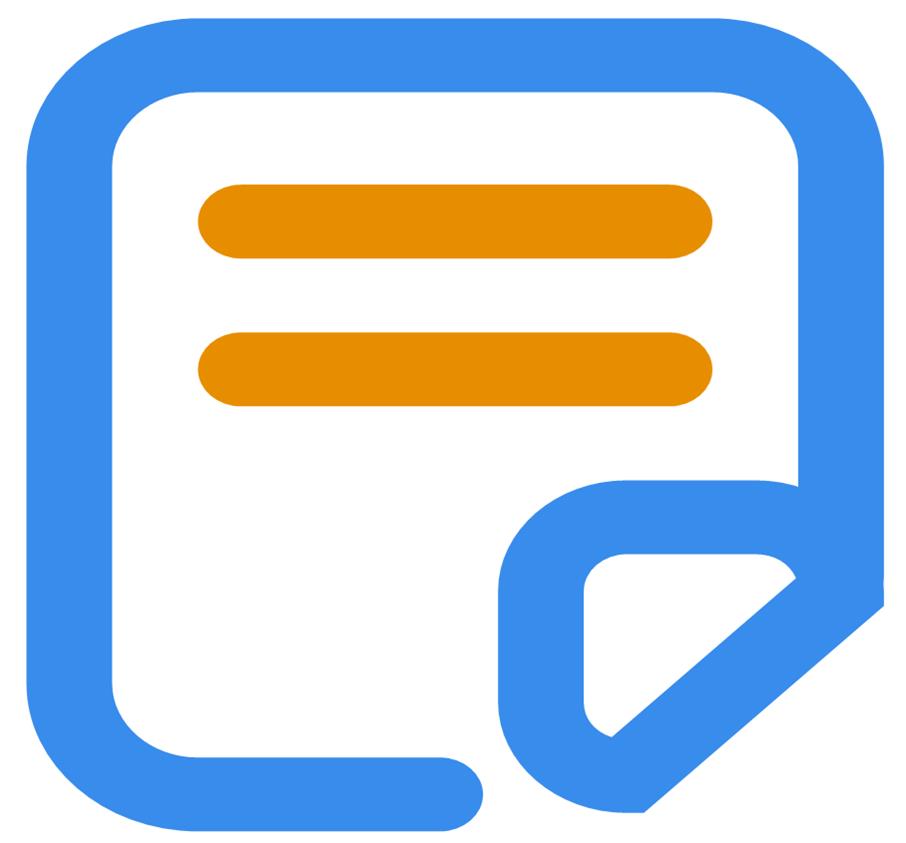}
			& \ding{56} & \ding{56}  \\   \hline
			ViT~\cite{dosovitskiy2020image}, Swin Transformer~\cite{liu2021swin}, MAE~\cite{he2022masked}  
			& \includegraphics[height=2.8mm, trim={0cm 3cm 0cm -4cm}]{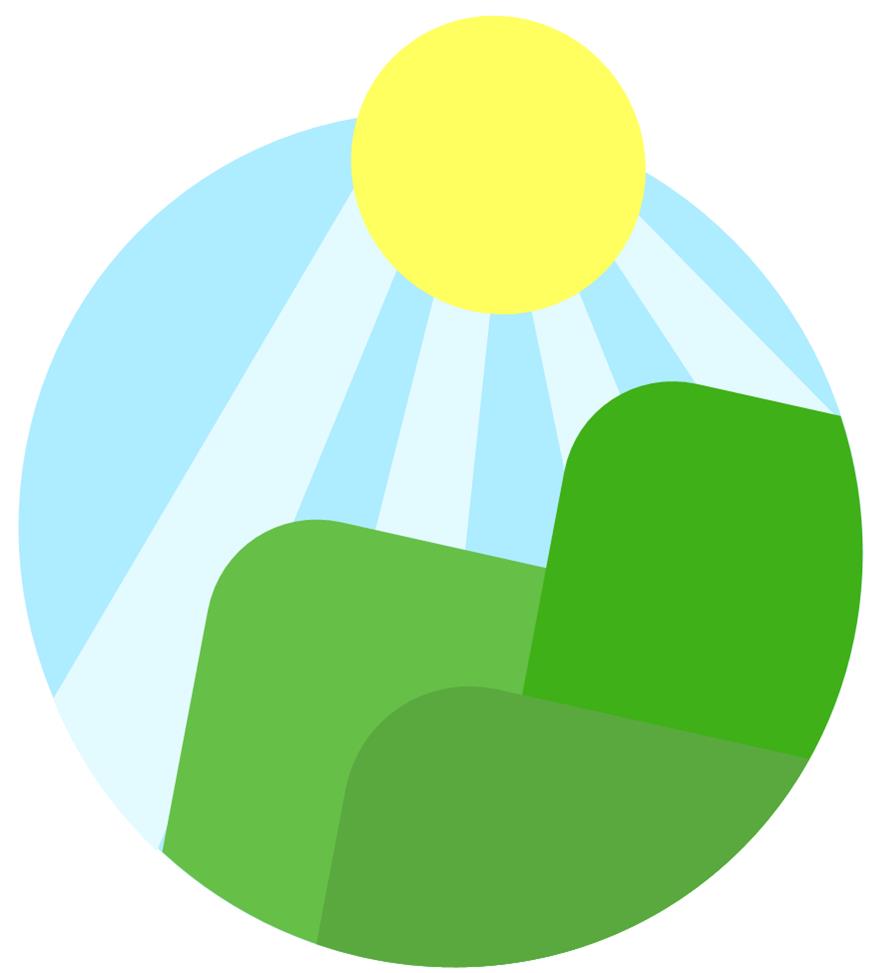}
			& \ding{56} & \ding{56}  \\  \hline
			Point Transformer\cite{zhao2021pointtransformer},~PCT~\cite{guo2021pct},~Point ViT~\cite{qian2022pix4point} 
			& \includegraphics[height=2.8mm, trim={0cm 3cm 0cm -4cm}]{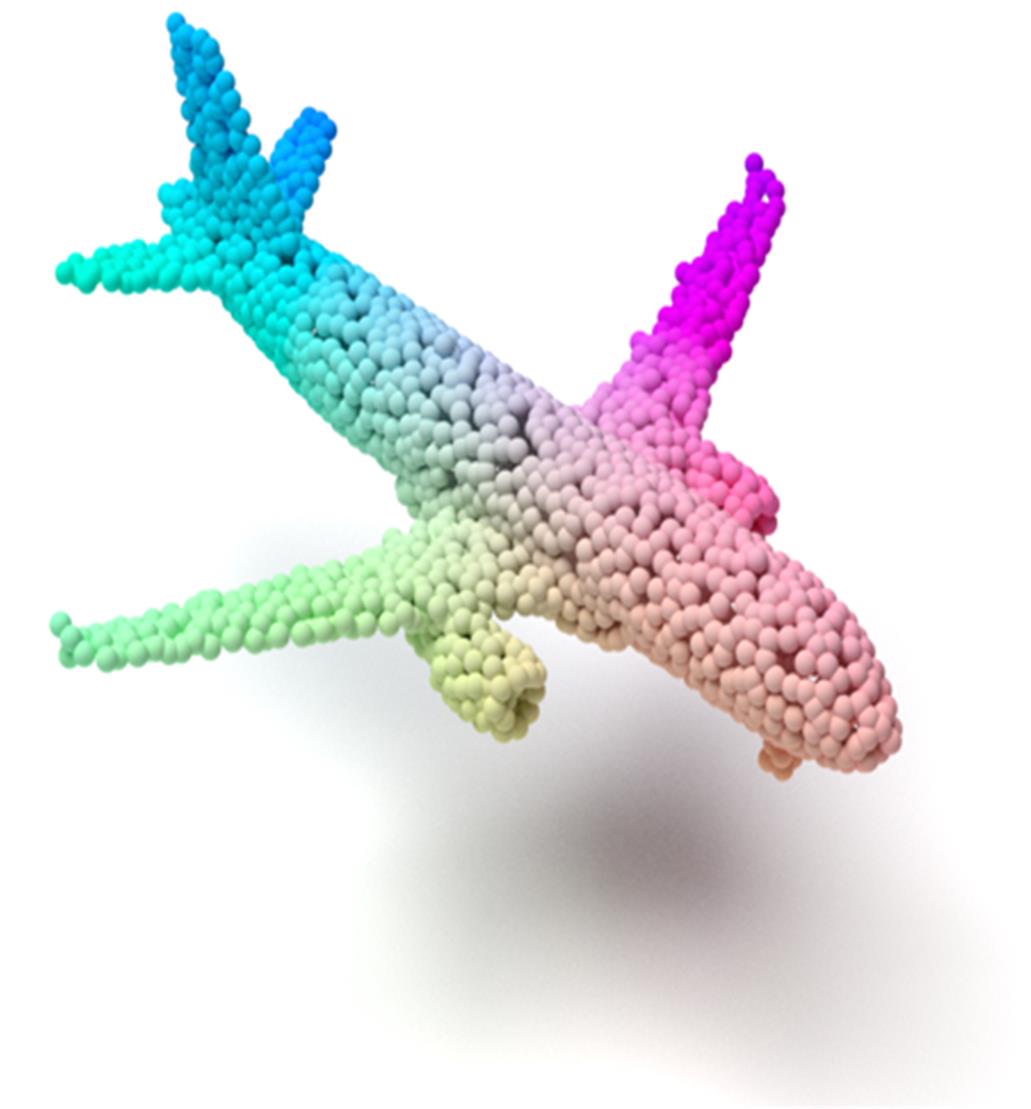}
			& \ding{56} & \ding{56}  \\\hline
			AST~\cite{gong2021ast},~SSAST~\cite{gong2022ssast}  
			& \includegraphics[height=2.8mm, trim={0cm 3cm 0cm -4cm}]{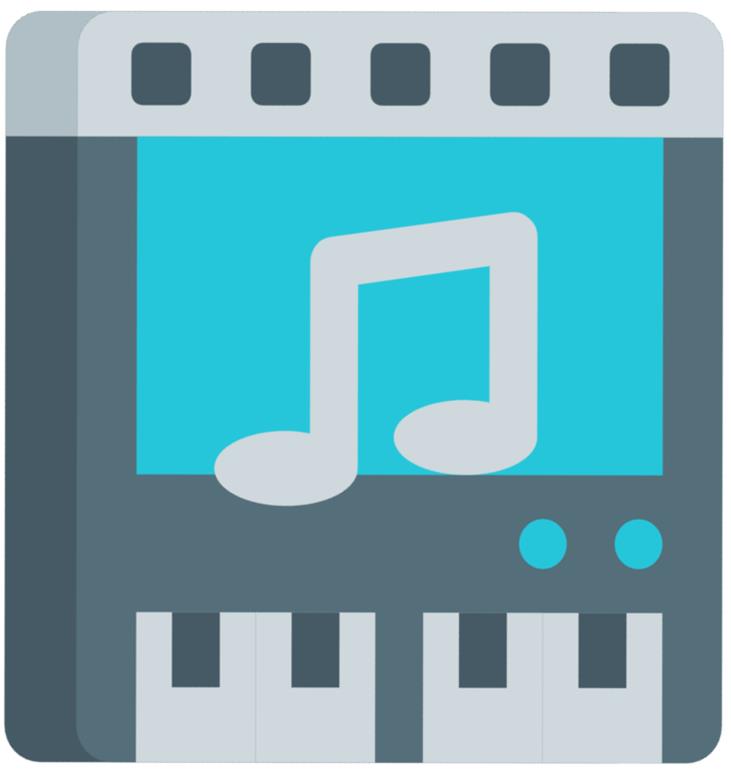}
			& \ding{56} & \ding{56} \\ \hline
			CLIP~\cite{radford2021learning},~Flamingo~\cite{alayrac2022flamingo},~VLMO~\cite{wang2021vlmo},~OFA~\cite{wang2022unifying} 
			& \includegraphics[height=2.8mm, trim={0cm 3cm 0cm -4cm}]{icons/text.jpg} \includegraphics[height=2.8mm, trim={0cm 3cm 0cm -4cm}]{icons/img.jpg}
			& \ding{56} & \ding{56} \\ \hline
			BEiT-3~\cite{wang2022image}  
			& \includegraphics[height=2.8mm, trim={0cm 3cm 0cm -4cm}]{icons/text.jpg} \includegraphics[height=2.8mm, trim={0cm 3cm 0cm -4cm}]{icons/img.jpg}
			& Several Layers & \ding{56} \\ \hline
			ImageBind~\cite{girdhar2023imagebind}  
			& \includegraphics[height=2.8mm, trim={0cm 3cm 0cm -4cm}]{icons/text.jpg} \includegraphics[height=2.8mm, trim={0cm 3cm 0cm -4cm}]{icons/img.jpg} \includegraphics[height=2.8mm, trim={0cm 3cm 0cm -4cm}]{icons/pcd.jpg} \includegraphics[height=2.8mm, trim={0cm 3cm 0cm -4cm}]{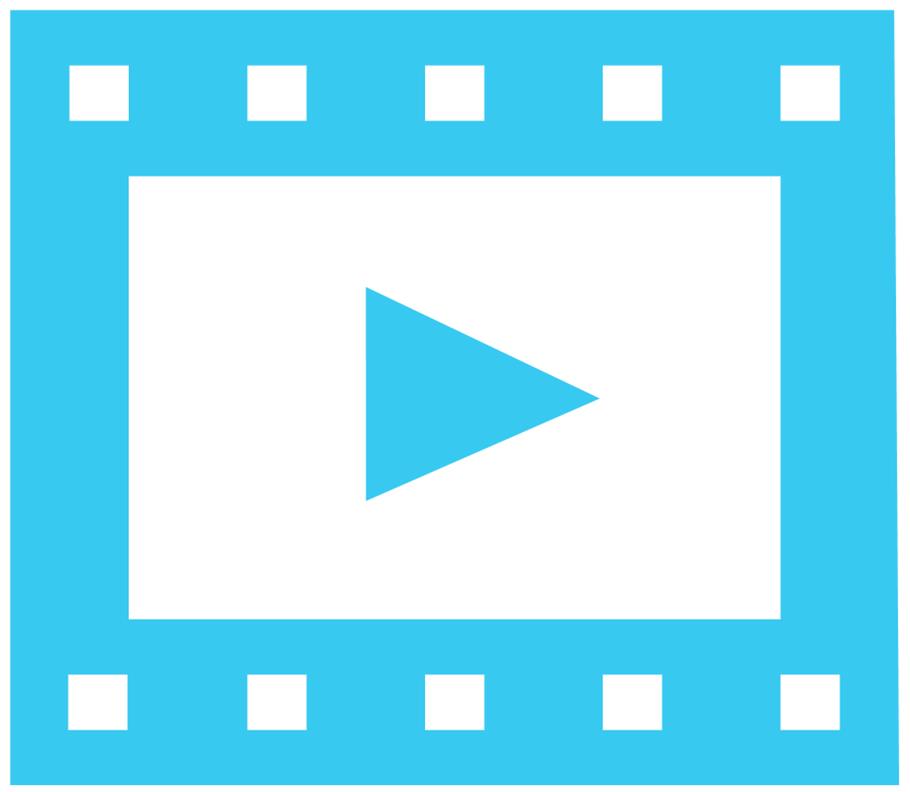} \includegraphics[height=2.8mm, trim={0cm 3cm 0cm -4cm}]{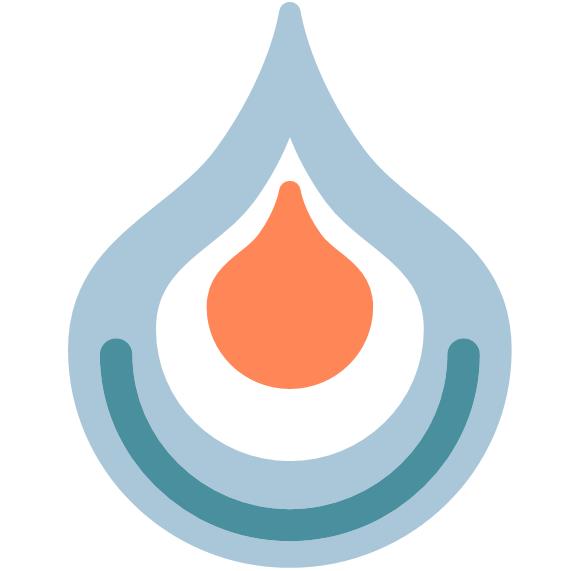} \includegraphics[height=2.8mm, trim={0cm 3cm 0cm -4cm}]{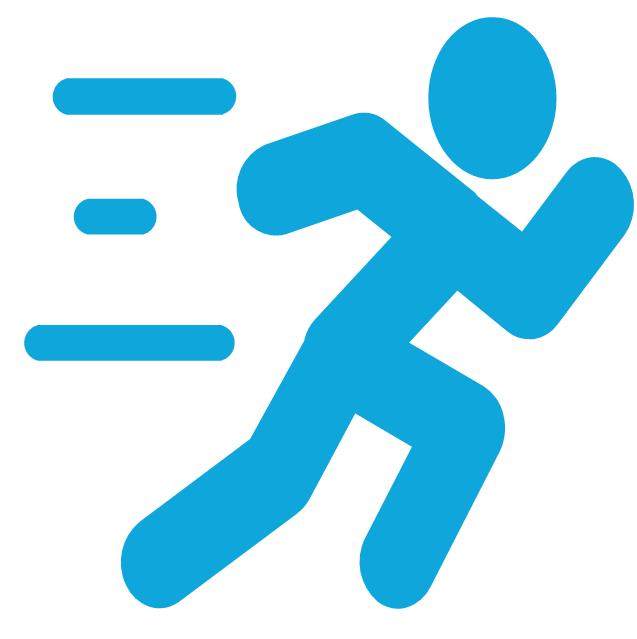} 
			& \ding{56} & \ding{56} \\ \hline
			Meta-Transformer~\pub{ours} 
			& \includegraphics[height=2.8mm, trim={0cm 3cm 0cm -4cm}]{icons/text.jpg} \includegraphics[height=2.8mm, trim={0cm 3cm 0cm -4cm}]{icons/img.jpg} \includegraphics[height=2.8mm, trim={0cm 3cm 0cm -4cm}]{icons/pcd.jpg} \includegraphics[height=2.8mm, trim={0cm 3cm 0cm -4cm}]{icons/audio.jpg} \includegraphics[height=2.8mm, trim={0cm 3cm 0cm -4cm}]{icons/video.jpg} \includegraphics[height=2.8mm, trim={0cm 3cm 0cm -4cm}]{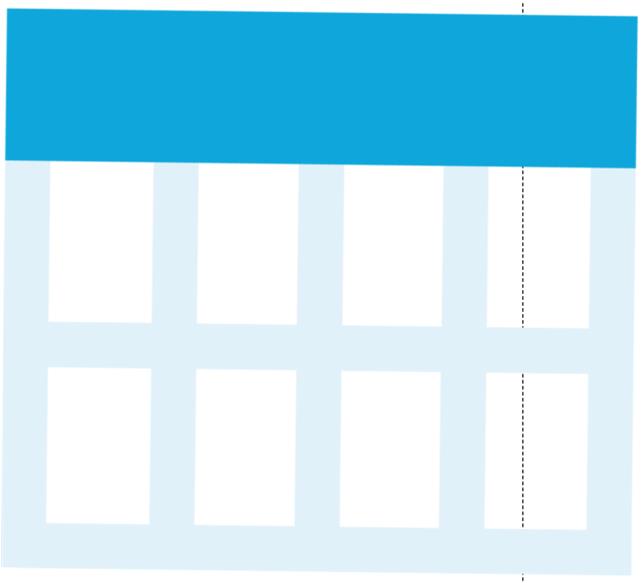} \includegraphics[height=2.8mm, trim={0cm 3cm 0cm -4cm}]{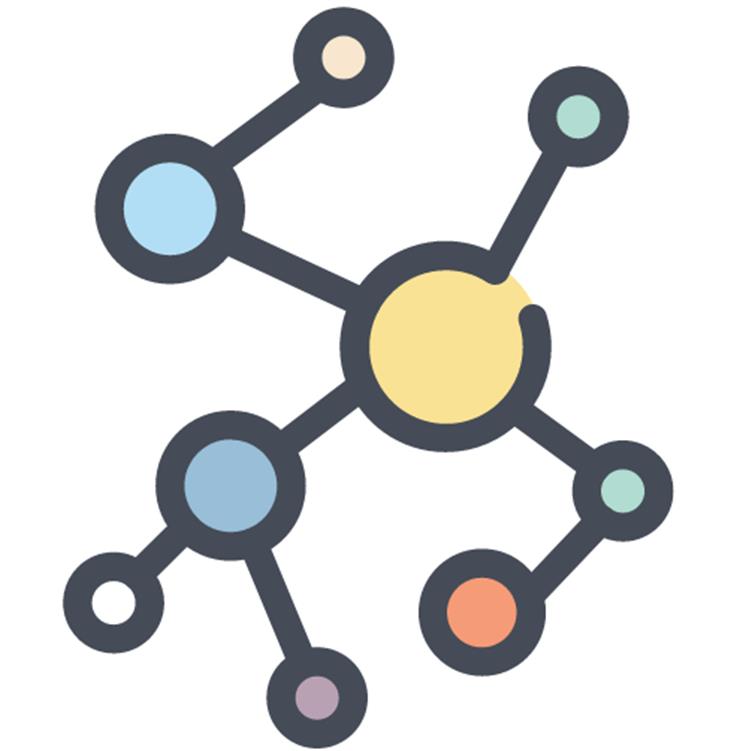} \includegraphics[height=2.8mm, trim={0cm 3cm 0cm -4cm}]{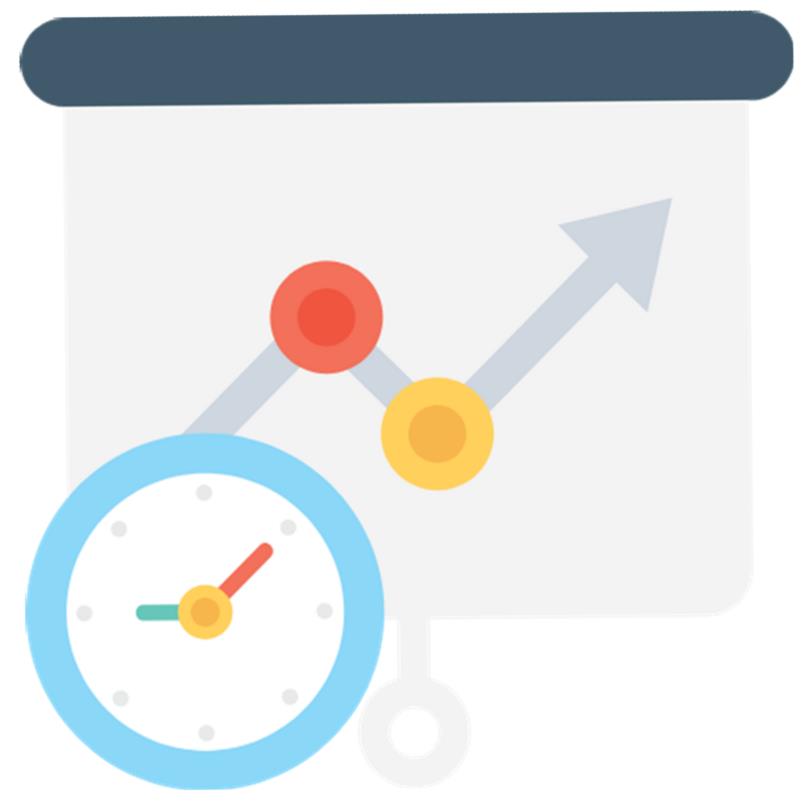} \includegraphics[height=2.8mm, trim={0cm 3cm 0cm -4cm}]{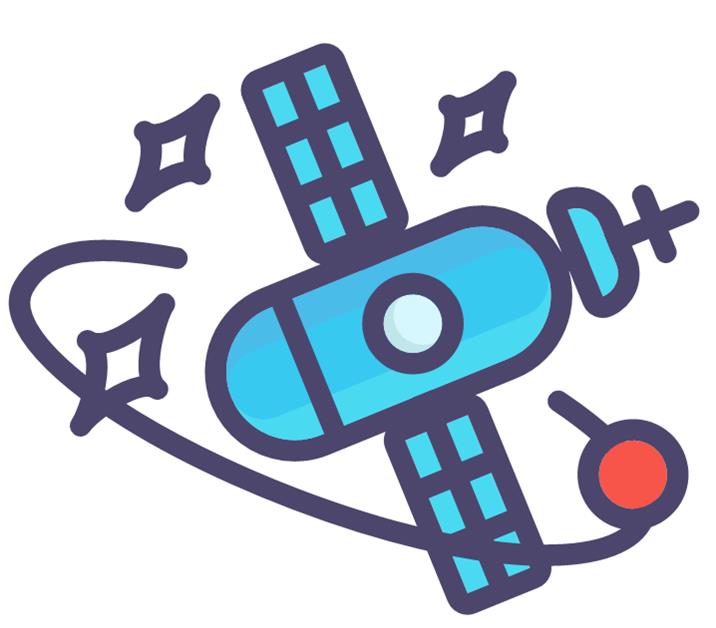} \includegraphics[height=2.8mm, trim={0cm 3cm 0cm -4cm}]{icons/imu.jpg} \includegraphics[height=2.8mm, trim={0cm 3cm 0cm -4cm}]{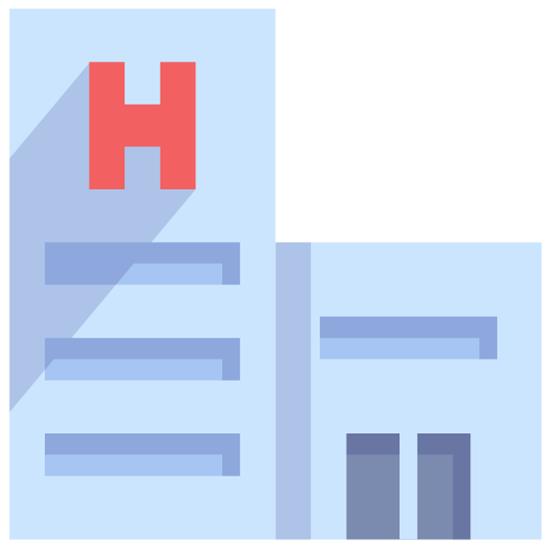} \includegraphics[height=2.8mm, trim={0cm 3cm 0cm -4cm}]{icons/infrared.jpg}& Whole Backbone & \ding{52} \\ \bottomrule
		\end{tabular}
	}   
	\label{tab:intrdo}
\end{table}

In this paper, We explore the potential of transformer architecture to process 12 modalities including images, natural language, point cloud, audio spectrogram, video, infrared, hyperspectral, X-Ray, IMU, tabular, graph, and time-series data, as shown in Figure~\ref{fig:teaser}. We discuss the learning process with transformers for each modality and address the challenges associated with unifying them into a single framework. Consequently, we propose a novel unified framework named Meta-Transformer for multimodal learning. \textbf{Meta-Transformer is the first framework to simultaneously encode data from a dozen of modalities using the same set of parameters}, allowing a more cohesive approach to multimodal learning (as shown in Table~\ref{tab:intrdo}). Meta-Transformer incorporates three simple and effective components: a modality-specialist (\S~\ref{sec:method:token}) for data-to-sequence tokenization, a modality-shared encoder (\S~\ref{sec:method:encoder}) for extracting representations across modalities, and task-specific heads for downstream tasks. Specifically, Meta-Transformer first transforms multimodal data into token sequences that share a common manifold space. Then, a modality-shared encoder with frozen parameters extracts representations, which are further adapted to individual tasks by updating the parameters of downstream task heads and lightweight tokenizers only. Finally, task-specific and modality-generic representations can be effectively learned by this simple framework.

We conduct extensive experiments on various benchmarks of 12 modalities.
By utilizing images of LAION-2B~\cite{radford2021learning} dataset for pretraining exclusively, Meta-Transformer demonstrates remarkable performance in processing data from multiple modalities, achieving consistently superior outcomes over state-of-the-art methodologies in different multimodal learning tasks. More detailed experimental settings can be found in \S~\ref{sec:details}.

\par
In conclusion, our contributions can be summarized as follows:
\begin{itemize}
	
	\item For multimodal research, we propose a novel framework, Meta-Transformer, which enables a unified encoder to simultaneously extract representations from multiple modalities with the same set of parameters.
	
	\item For multimodal network design, we comprehensively examine the functions of transformer components such as embeddings, tokenization, and encoders in processing various modalities. Meta-Transformer provides valuable insights and sparks a promising new direction in developing a modality-agnostic framework capable of unifying all modalities.
	
	\item Experimentally, Meta-Transformer achieves outstanding performance on various datasets regarding 12 modalities, which validates the further potential of Meta-Transformer for unified multimodal learning.
	
\end{itemize}
% --------------------------------------------------------------------------
\section{Related Work}~\label{sec:related}\vspace{-8mm}
\subsection{Single-Modality Perception}  ~\label{sec:related:single}
The development of various neural networks facilitates the perception of machine intelligence~\cite{mcculloch1943logical,hearst1998support,he2016deep,vaswani2017attention}.

%########### MLP  ####################################
\textbf{Multi-Layer Perceptron for pattern recognition.}
At the beginning, support vector machine (SVM) and multi-layer perceptron (MLP) are applied to text~\cite{xu2003representative}, image~\cite{lecun1989handwritten}, point cloud~\cite{qi2017pointnet}, and audio~\cite{dhanalakshmi2009classification} classification. These innovative works merit the feasibility of introducing AI to pattern recognition. \par
%########### CNN/ RNN variants #######################
\textbf{Recurrent \& Convolutional Neural Network.}
Hopfield Network~\cite{hopfield1982neural} is the original form of recurrent networks, then LSTM~\cite{hochreiter1997long} and GRU~\cite{chung2014empirical} further explore the advantages of RNNs in sequence modeling and application in NLP tasks~\cite{nallapati2016abstractive,cho2014learning,tang2015document}, which is also widely applied in audio synthesis~\cite{kalchbrenner2018efficient}.  Meanwhile, the success of CNNs  including LeNet~\cite{lecun1998gradient}, AlexNet~\cite{krizhevsky2017imagenet}, VGG~\cite{simonyan2014very}, GoogleNet~\cite{szegedy2015going} and ResNet~\cite{he2016deep} in image recognition greatly promote the application of CNNs in other fields such as text classification~\cite{zhang2015character,zhang2015sensitivity}, point cloud understanding~\cite{li2018pointcnn,maturana2015voxnet,thomas2019kpconv}, and speech classification~\cite{abdel2014convolutional}. 

%########### Transformer variants #######################
\textbf{Transformer.}
Recently, transformer architecture~\cite{vaswani2017attention} has been adopted in various tasks such as text understanding~\cite{devlin2018bert} and generation~\cite{brown2020language} in NLP, classification~\cite{dosovitskiy2020image}, detection~\cite{carion2020end} and segmentation~\cite{xie2021segformer} in images, point cloud understanding~\cite{guo2021pct,zhao2021pointtransformer}, and audio recognition~\cite{gong2021ast,gong2022ssast}. 

However, similar to applications of CNNs and RNNs, these networks are modified according to distinct properties of modalities. There is no common architecture for modality-agnostic learning. More importantly, information from different modalities can be complementary~\cite{liu2022bevfusion,zhang2020fusionnet,wang2022p2p}, it's significant to design a framework that can encode data from different modalities and bridge these complicated representations via a shared parameter space.
% --------------------------------------------------------------------------
\subsection{Transformed-based Multimodal Perception}  ~\label{sec:related:multi}
The advantages of transformers for perception are the global receptive field and similarity modeling, which prominently facilitate the development of multimodal perception.
MCAN~\cite{yu2019deep} proposes the deep modular co-attention networks between vision and language, which performs the cross-modal alignment by concisely maximizing the cross-attention. Then it becomes a consensus~\cite{wang2021vlmo,wang2021simvlm,wang2022unifying,wang2022image} to utilize a cross-attention mechanism to bridge different modalities. With the success of pretrain-finetune paradigm, more works are getting focused on how to effectively align representations extracted across modalities by pretraining. VL-BERT~\cite{su2019vl} pioneers modality-aligned representations for generic vision-language understanding with the MLM paradigm. Then Oscar~\cite{li2020oscar} described the object semantics in both visual and textural contents. Frameworks such as Vinvl~\cite{zhang2021vinvl}, Simvlm~\cite{wang2021simvlm}, VLMO~\cite{wang2021vlmo}, ALBEF~\cite{li2021align}, and Florence~\cite{yuan2021florence} further explore the advantages of joint representations across vision-language modalities in terms of semantic consistency. 

Multimodal models are also utilized for few-shot learning~\cite{alayrac2022flamingo}, sequence-to-sequence learning~\cite{wang2022unifying}, contrastive learning~\cite{yu2022coca}. BEiT-v3~\cite{wang2022image} proposes to take images as a foreign language with a more fine-grained cross-modal mask-and-reconstruction process, sharing partial parameters. And MoMo~\cite{chada2023momo} further explores the training strategy and objective functions while using the same encoder for images and texts. 

Despite these advances, there remain significant obstacles to designing unified multimodal networks due to differences between modalities. Additionally, most research in this area has focused on vision and language tasks, and may not directly contribute to challenges such as 3D point cloud understanding, audio recognition, or other modalities. The Flamingo model~\cite{alayrac2022flamingo} represents a powerful few-shot learner, but its transferability to point clouds is limited, and it remains a challenge to leverage prior knowledge from one modality to benefit the others. In other means, existing multimodal methods have limited extensibility on more modalities, although they have taken expensive training costs. Addressing these discrepancies is dependent on bridging different modalities using the same set of parameters, akin to how a bridge connects multiple river banks.
% --------------------------------------------------------------------------
\section{Meta-Transformer}~\label{sec:method}
\begin{figure*}[t]
	\centering
	\includegraphics[width=0.98\columnwidth]{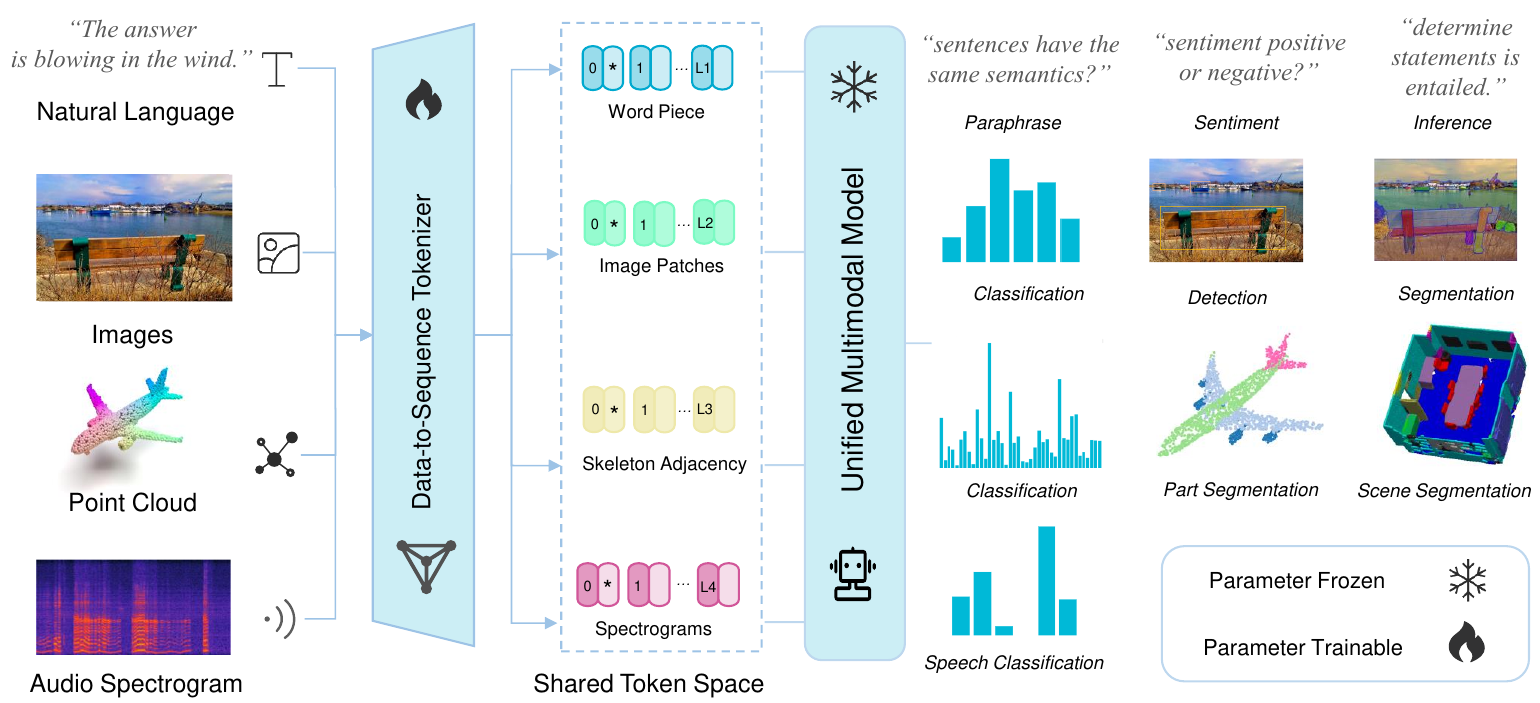}
	\caption{ Meta-Transformer consists of data-to-sequence tokenization, unified feature encoding, and down-stream task learning. The framework is illustrated with text, image, point cloud, and audio.
	}
	\label{fig:framework}
\end{figure*}

In this section, we depict the proposed framework, Meta-Transformer, in detail. Meta-Transformer unifies the multiple pipelines of processing data from different modalities and fulfills encoding texts, images, point clouds, audio, and the other 8 modalities with a shared encoder. To achieve this, Meta-Transformer is composed of a data-to-sequence tokenizer to project data to a shared embedding space, a modality-agnostic encoder to encode the embedding of different modalities, and task-specific heads to perform downstream predictions, as shown in Fig.~\ref{fig:framework}.

%-------------------------------------------
%\vspace{-0.15in}
% --------------------------------------------------------------------------
\subsection{Preliminary}~\label{sec:method:pre}
% --------------------------------------------------------------------------
Formally, we denote the input space of $n$ modalities as $\{\mathcal{X}_1, \mathcal{X}_2, \cdots, \mathcal{X}_n\}$, while $\{\mathcal{Y}_1, \mathcal{Y}_2, \cdots, \mathcal{Y}_n\}$ are the corresponding label spaces. In addition, we assume there exists an \textbf{effective} parameter space $\Theta_i$ for each modality, where any parameter $\theta_i \in \Theta_i$ can be utilized for processing data $\boldsymbol{x}_i \in \mathcal{X}_i$ from that modality. We say that the essence of Meta-Transformer is to find a shared $\theta^{*}$ that satisfies:
\begin{equation}
	\begin{aligned}
		\theta^{*} \in \Theta_1 \cap \Theta_2 \cap \Theta_3 \cap \cdots \Theta_n,
		\label{eq:pre:params}
	\end{aligned}
\end{equation} with the hypothesis:
\begin{align}
	\Theta_1 \cap \Theta_2 \cap \Theta_3 \cap \cdots \Theta_n \neq \varnothing.
\end{align}
The multimodal neural networks can be formulated as a unified mapping function $\mathcal{F}: \boldsymbol{x}\in \mathcal{X} \rightarrow \hat{y} \in \mathcal{Y}$, where $\bm{x}$ is the input data coming from any modality $\{\mathcal{X}_1, \mathcal{X}_2, \cdots, \mathcal{X}_n\}$ and $\hat{y}$ denotes the prediction of the network. Let's denote $y$ as the ground truth labels, the multimodal pipeline can be formulated as:
\begin{equation}
	\begin{aligned}
		\hat{y} = \mathcal{F}(\boldsymbol{x};\theta^{*}), 
		\hspace{1.5mm}
		\theta^* = \underset{x \in \mathcal{X}}{ \arg \min}[\mathcal{L}(\hat{y},y)].
	\end{aligned}
	\label{eq:pre:target}
\end{equation}

% --------------------------------------------------------------------------
\subsection{Data-to-Sequence Tokenization}~\label{sec:method:token}
% --------------------------------------------------------------------------
We propose a novel meta-tokenization scheme designed to transform data across various modalities into token embeddings, all within a shared manifold space. This approach is then applied to tokenization, taking into account the practical characteristics of modality, as illustrated in Figure~\ref{fig:token}. We take text, images, point clouds, and audio as examples. More details can be found in supplementary materials. In specific, we use $\boldsymbol{x}_{T}$, $\boldsymbol{x}_{I}$, $\boldsymbol{x}_{P}$, and $\boldsymbol{x}_{A}$ to denote a data sample of text, image, point cloud, and audio spectrogram.

\textbf{Natural Language}.
Following the common practice~\cite{devlin2018bert,liu2019roberta}, we use WordPiece embeddings~\cite{wu2016google} with a 30,000 token vocabulary. WordPiece segments original words into subwords. For example, the original sentence: ``The supermarket is hosting a sale'', could be converted by WordPiece to: ``\_The \_super market \_is \_host ing \_a \_sale''.

In this case, the word ``supermarket'' is divided into two subwords ``\_super'' and ``market'' and the word ``hosting'' is divided into ``\_host'' and ``ing'', while the rest words are unchanged and still single units. The front of the first character of each original word will be stacked with a special character ``\_'', indicating the beginning of a natural word. Each subword is corresponding to a unique token in a vocabulary, then is projected to a high-dimensional feature space with word embedding layers. As a result, each input text is transformed to a set of token embeddings $\bm{x} \in \mathbb{R}^{n\times D}$, where $n$ is the number of tokens and $D$ is the dimension of embedding.

%-------------------------------------------
%\vspace{-0.15in}
\begin{figure*}[t]
	\centering
	\includegraphics[width=0.98\columnwidth]{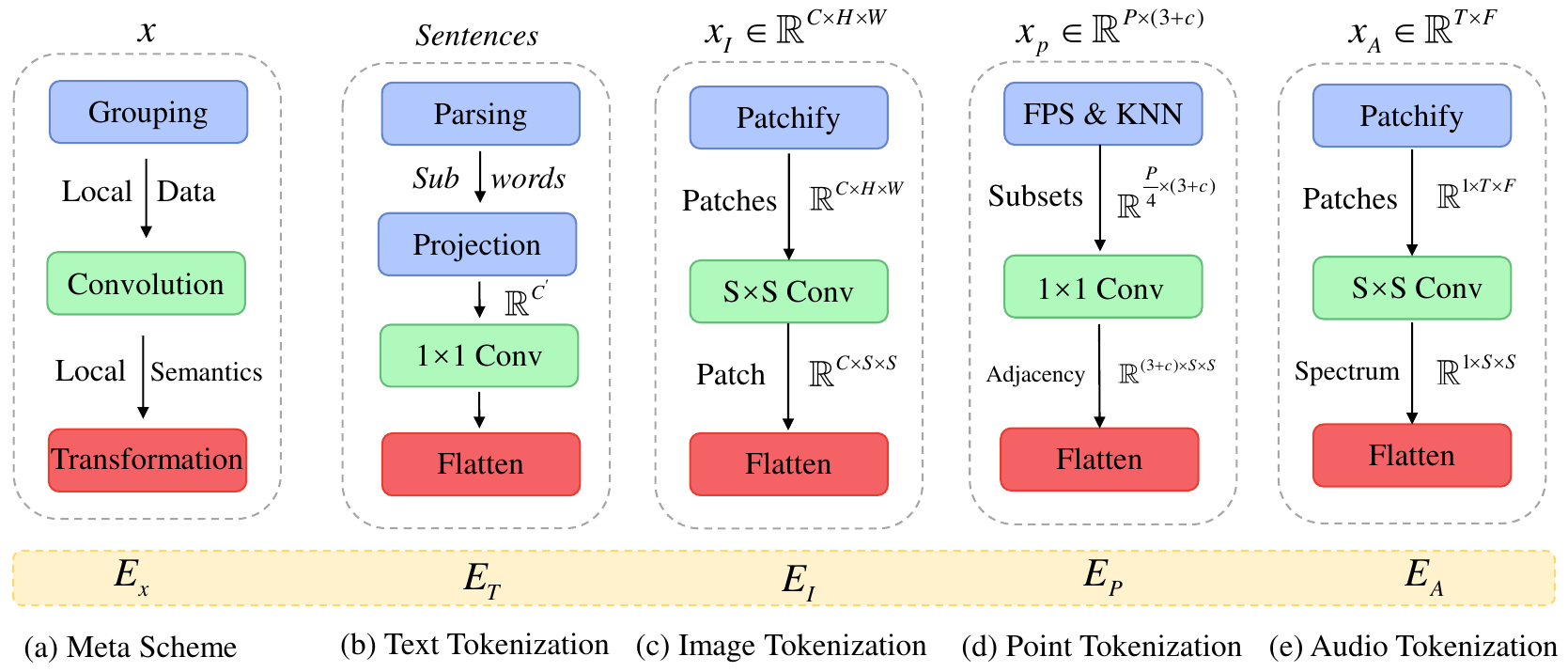}
	\caption{ Illustration of Data-to-Sequence Tokenization~\ref{sec:method:token}. We propose the meta scheme in (a) containing grouping, convolution, and transformation progress. Then (b)-(e) represents the building blocks applied with our meta scheme on texts, images, point clouds, and audio spectrograms.}
	\label{fig:token}
\end{figure*}
% --------------------------------------------------------------------------
% --------------------------------------------------------------------------
\textbf{Images}. 
% Generally, transformer deals with input as the sequence of token embeddings. 
To accommodate 2D images, we reshape the image $\boldsymbol{x} \in \mathbb{R}^{H \times W \times C}$ into a sequence of flattened 2D patches $\boldsymbol{x}_p \in \mathbb{R}^{N_s \times (S^2 \cdot C)}$, where $(H, W)$ represents the original image resolution, $C$ denotes the number of channels; $S$ is the patch size, and $N_s=(HW/S^2)$ is the resulting number of patches. After that, a projection layer is utilized to project the embedding dimension to $D$:
\begin{equation}
	\boldsymbol{x}_I \in \mathbb{R}^{ C \times H \times W } \rightarrow \boldsymbol{x}_I^\prime \in \mathbb{R}^{N_s \times (S^2 \cdot C)} \rightarrow \boldsymbol{x}_I^{\prime\prime} \in \mathbb{R}^{N_s \times D }.
	\label{eq:image:token}
\end{equation}
Note that we use the same operation for infrared images but the linear projection for hyperspectral images. In addition, we simply replace 2D convolution layers with 3D convolution for video recognition. More details can be found in \ref{sec:single-modal:video} and \ref{sec:single-modal:hyper}.
\par
% --------------------------------------------------------------------------
\textbf{Point Cloud}.
To learn 3D patterns with transformers, we convert point clouds from raw input space to the token embedding space. $\mathcal{X}=\{\boldsymbol{x}_i\}_{i=1}^P$ denotes a point cloud of $P$ points, where $\boldsymbol{x}_i = (\boldsymbol{p}_i,\boldsymbol{f}_i)$, $\boldsymbol{p}_i\in \mathbb{R}^3$ represents the 3D coordinates, and $\boldsymbol{f}_i\in \mathbb{R}^c$ is feature of the $i$-th point. Generally, $\boldsymbol{f}_i$ contains visual hints such as color, viewpoint, normal, etc. We employ the Farthest Point Sampling ($\mathtt{FPS}$) operation to sample a representative skeleton of original point clouds with a fixed sampling ratio (1/4). Then we employ $K$-Nearest Neighbor ($\mathtt{KNN}$) to group neighboring points. Based on grouped sets containing local geometric prior, we construct the adjacency matrix with center points of grouped subsets to further undercover the comprehensive structural information of 3D objects and 3D scenes. Finally, we aggregate the structural representations from $K$ subsets. We obtain point embeddings as:
\begin{equation}
	\boldsymbol{x}_P \in \mathbb{R}^{ P \times (3+c)} \rightarrow \boldsymbol{x}_P^\prime \in \mathbb{R}^{\frac{P}{4} \times \frac{D}{2}} \rightarrow \boldsymbol{x}_P^{\prime\prime} \in \mathbb{R}^{ \frac{P}{16} \times D }.
	\label{eq:pcd:token}
\end{equation}

% --------------------------------------------------------------------------
\textbf{Audio Spectrogram}.
% --------------------------------------------------------------------------
Initially, we pre-process the audio waveform with the duration of $t$ seconds with log Mel filterbank~\cite{schneider2019wav2vec}. Then we employ the Hamming window with a stride of $t_s$ on the frequency of $f_s$ to split the original wave into $l = (t / t_s) $ intervals and further transform the original wave into $l$-dimensional filterbank. 

Subsequently, we split the spectrogram into patches from time and frequency dimensions with the same patch size of $S$. Different from image patches, audio patches overlap on spectrograms. Following AST~\cite{gong2021ast}, we also choose to split whole spectrograms into $N_s = 12[(100t - 16)/10]$ patches by $S\times S$ convolution, then we flatten patches into token sequences. Finally, we summarize the process:
\begin{equation}
	\boldsymbol{x}_A \in \mathbb{R}^{ T \times F} \rightarrow \boldsymbol{x}_A^\prime \in \mathbb{R}^{  N_s \times S \times S } \rightarrow \boldsymbol{x}_A^{\prime\prime} \in \mathbb{R}^{ (N_s \cdot D/S^2) \times D },
	\label{eq:audio:token}
\end{equation}
where $T$ and $F$ denote time and frequency dimensions.

\subsection{Unified Encoder}~\label{sec:method:encoder}
% --------------------------------------------------------------------------
After transforming the raw inputs to token embedding space, we leverage a unified transformer encoder with frozen parameters to encode the sequences of token embeddings from different modalities.

\textbf{Pretraining}.  We utilize ViT~\cite{dosovitskiy2020image} as the backbone network and pre-train it on the LAION-2B dataset with contrastive learning, which reinforces the ability for generic token encoding. After pretraining, we freeze the parameters of the backbone network. In addition, for text understanding, we utilize the pretrained text tokenizer of CLIP~\cite{radford2021learning} to segment sentences into subwords and transform subwords into word embeddings.\par
% --------------------------------------------------------------------------

\textbf{Modality-Agnostic Learning}. 
Following common practice~\cite{devlin2018bert,dosovitskiy2020image}, we prepend a learnable token $x_{\texttt{CLS}}$ to the sequence of token embeddings, and the final hidden state of $x_{\texttt{CLS}}$ token ($\boldsymbol{z}_L^0$) serves as the summary representation of the input sequence, which is usually utilized for performing recognition. 

To reinforce positional information, we incorporate position embeddings into the token embeddings. Recall that we tokenize the input data to 1D embeddings, thus, we opt for standard learnable 1D position embeddings. In addition, we do not observe substantial performance improvements using more sophisticated 2D-aware position embeddings on image recognition. We simply fuse the position embeddings and the content embeddings with an element-wise addition operation, and the resulting embedding sequences are then fed into the encoder.

The transformer encoder with a depth of $L$ compromises multiple stacked multi-head self-attention (MSA) layers and MLP blocks. The input token embeddings are fed into an MSA layer first, then an MLP block. Then the output of $(\ell - 1)$-{th} MLP block serves as the input of $\ell$-th MSA layer. Layer Normalization (\texttt{LN}) is appended before each layer and the residual connection is applied after each layer. The MLP contains two linear FC layers along with a \texttt{GELU} non-linear activation. The formulation of the transformer is:
\begin{align}
	\boldsymbol{z}_0 &= [ \boldsymbol{x}_\texttt{CLS}; \,  \boldsymbol{E}_{\boldsymbol{x}_1}; \,  \boldsymbol{E}_{\boldsymbol{x}_2}; \cdots; \,  \boldsymbol{E}_{\boldsymbol{x}_n} ] + \boldsymbol{E}_{pos},
	&& \boldsymbol{E} \in \mathbb{R}^{n \times D},\, \boldsymbol{E}_{pos}  \in \mathbb{R}^{(n + 1) \times D} \label{eq:embedding} \\
	\boldsymbol{z^\prime}_\ell &= \operatorname{MSA}(\mathtt{LN}(\boldsymbol{z}_{\ell-1})) + \boldsymbol{z}_{\ell-1}, && \ell=1\ldots L \label{eq:msa_apply} \\
	\boldsymbol{z}_\ell &= \operatorname{MLP}(\mathtt{LN}(\boldsymbol{z^\prime}_{\ell})) + \boldsymbol{z^\prime}_{\ell}, && \ell=1\ldots L  \label{eq:mlp_apply} \\
	\boldsymbol{y} &= \mathtt{LN}(\boldsymbol{z}_L^0) \label{eq:final_rep} 
\end{align}where $\boldsymbol{E}_x$ denotes the token embeddings from proposed tokenizer and $n$ denotes the number of tokens. 
We augment patch embeddings and learnable embedding with position embeddings $\bm{E}_{pos}$.

\subsection{Task-Specific Heads}
After obtaining learning representations, we feed representations to the task-specific heads $h(\cdot;\theta_h)$, which consists mainly of MLPs and varies from modalities and tasks. The learning objective of Meta-Transformer can be summarized as:
\begin{equation}
	\label{eq:heads}
	\hat{\boldsymbol{y}} = \mathcal{F}(\boldsymbol{x};\theta^{*}) = h \circ g \circ f (\boldsymbol{x}), 
	\hspace{2.5mm}
	\theta^{*} = \mathop{\arg \min}_{\theta} \mathcal{L}(\hat{y}, y),
\end{equation}
where $f(\cdot)$, $g(\cdot)$, and $h(\cdot)$ denote the function of tokenizer, backbone, and heads, respectively.
% --------------------------------------------------------------------------
\section{Experiments}~\label{sec:exp}
In this section, we perform experiments on each of the 12 modalities. We demonstrate the potential of Meta-Transformer for multimodal perception. A summary of our experimental design is shown in Table~\ref{tab:summary} and more experimental details can be found in \S~\ref{sec:multi-modal:audio-image}.
\noindent\subsection{Experimental Setups} ~\label{sec:exp:setup}
\textbf{Text understanding}. For text understanding evaluation, we employ the General Language Understanding Evaluation (GLUE) benchmark~\cite{wang2018glue} which incorporates several different datasets, covering a wide range of natural language understanding tasks.

\textbf{Image understanding}. 1) Classification: we conduct experiments on ImageNet-1K~\cite{deng2009imagenet} which contains approximately 1.3 million images with 1000 categories. Following common practices~\cite{wang2021pyramid,liu2021swin,liu2022convnet}, base-scale models are trained for 300 epochs, while large models are pre-trained on ImageNet-22K (14.2 million images) for 90 epochs and fine-tuned on ImageNet-1K for another 20 epochs. 2) Object Detection: we conduct experiments on the MS COCO dataset~\cite{lin2014microsoft} using Mask R-CNN~\cite{he2017mask} as the detector and training each model for 12 epochs. 3) Semantic Segmentation: we train the segmentation head UperNet~\cite{xiao2018unified} on ADE20K~\cite{zhou2017scene} for 160k iterations, providing a fair comparison with previous CNN-based and transformer-based backbones.

\textbf{Infrared, X-Ray, and Hyperspectral data understanding}. We conduct experiments on infrared image, X-Ray scan, and hyperspectral data recognition with RegDB~\cite{nguyen2017person}, Chest X-Ray~\cite{rahman2020reliable}, and Indian Pine~\footnote{\url{https://github.com/danfenghong/IEEE_TGRS_SpectralFormer/blob/main/data/IndianPine.mat}} datasets, respectively.

\textbf{Point cloud understanding}. 1) Classification: to assess the performance of Meta-Transformer in 3D object classification, we use the ModelNet-40~\cite{wu2015modelnet} benchmark, consisting of CAD models across 40 classes, with 9,843 training samples and 2,468 validation samples. 2) Semantic segmentation: to evaluate performance in 3D point cloud segmentation, we assess the model on both S3DIS~\cite{armeni2016s3dis} and ShapeNetPart~\cite{yi2016scalable} datasets. The S3DIS dataset encompasses 6 large indoor areas and 13 semantic classes, comprising 271 rooms. The ShapeNetPart dataset includes 16,880 object models across 16 shape categories.

\textbf{Audio recognition}. For audio recognition, we utilize the Speech Commands V2~\cite{warden2018speech} dataset, which consists of 105,829 one-second recordings of 35 common speech commands.

\textbf{Video recognition}. For video understanding, we conduct experiments on the UCF101~\cite{ucf} dataset for action recognition, with more details presented in~\S~\ref{sec:single-modal:video}.

\textbf{Time-series forecasting}. For time-series forecasting, we conduct experiments on ETTh1~\cite{haoyietal-informer-2021}, Traffic\footnote{\href{https://pems.dot.ca.gov/}{https://pems.dot.ca.gov/}}, Weather\footnote{\href{https://www.bgc-jena.mpg.de/wetter/}{https://www.bgc-jena.mpg.de/wetter/}}, and Exchange~\cite{lai2018modeling} datasets. We use the tokenizer of Autoformer~\cite{wu2021autoformer}.

\textbf{Graph understanding}. We conduct experiments on the PCQM4M-LSC dataset~\cite{hu2021ogb}, which is a large-scale dataset consisting of 4.4 million organic molecules with up to 23 heavy atoms with their corresponding quantum-mechanical properties. With the target of predicting molecular properties using machine learning, it has plenty of applications in drug discovery, and material science. 

\textbf{Tabular analysis}. We conduct experiments on adult and bank marketing from UCI repository~\footnote{\href{http://archive.ics.uci.edu/ml/}{http://archive.ics.uci.edu/ml/}}. We use the tokenizer of TabTransformer~\cite{huang2020tabtransformer} to encode raw tabular data.

\textbf{IMU recognition}. To evaluate the ability of Meta-Transformer to understand the inertial motion systems, we conduct experiments of IMU sensor classification on the Ego4D~\cite{grauman2022ego4d} dataset.

\begin{table}[ht]
	\centering
	\caption{Summary of experimental settings across different modalities. We report the task, dataset, and data scale for each modality.}
	\begin{tabular}{l|c| c | c}
		\toprule 
		Modalities  &  Tasks & Datasets & Data Scale \\ \hline
		\includegraphics[height=2.6mm, trim={0cm 3cm 0cm -4cm}]{icons/text.jpg} Text & Classification & GLUE Benchmark & 330K \\ \hline
		\multirow{3}{*}{ \includegraphics[height=3.0mm, trim={6cm 1cm 1cm 1cm}]{icons/img.jpg} Image} & Classification & ImageNet-1K & 1.3M  \\
		& Detection & MS COCO & 118K\\ 
		& Segmentation & ADE-20K & 20K\\ \hline
		\multirow{3}{*}{\includegraphics[height=3.0mm]{icons/pcd.jpg} Point Cloud} & Shape Classification & ModelNet-40 & 9K \\
		& Scene Segmentation & S3DIS & 400M Points\\
		& Object Segmentation & ShapeNetPart & 16K\\ \hline
		\includegraphics[height=2.8mm, trim={0cm 3cm 0cm -4cm}]{icons/audio.jpg} Audio & Classification & Speech commands v2 & 105K\\ \hline
		\includegraphics[height=2.4mm, trim={0cm 3cm 0cm -4cm}]{icons/video.jpg} Video & Action Recognition & UCF101 & 14K\\
		\hline
		\includegraphics[height=2.8mm, trim={0cm 3cm 0cm -4cm}]{icons/infrared.jpg} Infrared & Classification & RegDB & 40K\\ \hline
		\includegraphics[height=2.5mm, trim={0cm 3cm 0cm -4cm}]{icons/hyper.jpg} Hyper-spectrum & Classification & Indian Pine & 10K\\ \hline
		\includegraphics[height=2.8mm, trim={0cm 3cm 0cm -4cm}]{icons/xray.jpg} X-Ray & Classification & Chest X-Ray & 112K \\ \hline
		\includegraphics[height=2.8mm, trim={0cm 3cm 0cm -4cm}]{icons/imu.jpg} IMU & Classification & Ego4D & 193K \\
		\hline
		\includegraphics[height=2.5mm, trim={0cm 3cm 0cm -4cm}]{icons/table.jpg} Tabular data & Prediction & Adult \& Bank & 32K-45K \\ \hline
		\includegraphics[height=2.8mm, trim={0cm 3cm 0cm -4cm}]{icons/graph.jpg} Graph data & Prediction & PCQM4M-LSC & 47M \\ \hline
		\includegraphics[height=2.8mm, trim={0cm 3cm 0cm -4cm}]{icons/time.jpg} Time-series & Forecasting & Exchange, Traffic, \textit{etc} & 5-36K \\
		\bottomrule
	\end{tabular}
	\label{tab:summary}
\end{table}

\textbf{Settings of Networks}: We follow the default settings of ViT~\cite{dosovitskiy2020image}. $\text{Meta-Transformer-B16}_F$ denotes Meta-Transformer with a base-scale encoder which contains 12 transformer blocks and 12 attention heads, and the image patch size is 16. For the base-scale encoder, the embedding dimension is 768 and the output dimension of MLP is 3,072.  `F' and `T' denotes that parameters of the encoder are \textit{Frozen} and further \textit{Tuned}, respectively.
\begin{table*}[ht]
	\centering
	\caption{\textbf{Experimental results for text understanding on the GLUE benchmark.} We compare existing advanced methods from paraphrasing, sentiment, duplication, inference, and answering tasks, and we report the pre-training settings and performances.}
	\label{tab:nlp}
	\resizebox{1.0\linewidth}{!}{
		\begin{tabular}{ l | c|c|c | ccccc }
			\toprule
			\multirow{3}{*}{Method} & \multicolumn{3}{c|}{Pretraining Settings} & \multicolumn{5}{c}{GLUE Benchmark} \\ 
			\cline{ 2 - 7 } \cline{6 - 9 } & \multirow{2}{*}{Modality} &  \multirow{2}{*}{Data} & \multirow{2}{*}{Size} &   SST-2	& MRPC	& QQP	& MNLI	& QNLI \\ 
			&  & & & Sentiment & Paraphrase & Duplication &  Inference & Answering \\ \hline
			BiLSTM+ELMo+Attn & - & - & - & 90.4 & 84.9 & 64.8 & 76.4 & 79.8 \\ \hline
			OpenAI GPT~\cite{radford2018improving} & \multirow{4}{*}{Language} & Book & 0.8B & 91.3 & 82.3  & 70.3 & 82.1 &87.4  \\ \cline{3-4}
			
			$\text{BERT}_{\text{BASE}}$~\cite{devlin2018bert} &   & \multirow{2}{*}{Wiki+Book} & \multirow{2}{*}{3.3B}  & {88.0} & {88.9} & 71.2 & {84.6} & {90.5} \\ 
			
			$\text{RoBERTa}_{\text{BASE}}$~\cite{liu2019roberta} &   &  &   
			& \textbf{96.0} & \textbf{90.0} & \textbf{84.0} & 84.0 & \textbf{92.0}  \\ 
			
			$\text{ChatGPT}$ &   & Various  & 4,5000B  
			& 92.0 & 66.0 & 78.0 & \textbf{89.3} & 84.0  \\ \hline
			
			$\text{Meta-Transformer-B16}_\text{F}$~\pub{ours} & \multirow{2}{*}{Image}  & \multirow{2}{*}{LAION-2B~\cite{radford2021learning}}  & \multirow{2}{*}{2B}  
			& 54.6 & 81.1 & 66.0 & 63.4 & 56.3  \\
			
			$\text{Meta-Transformer-B16}_\text{T}$~\pub{ours} &   &   &   
			& 81.3 & 81.8 & 78.0 & 70.0 & 60.3  \\
			\bottomrule
		\end{tabular}
	}
\end{table*}
%----------------------------------------------------------------------------
\begin{table*}[t]
	\centering
	\caption{\textbf{Experimental results for image understanding}. We conduct experiments in classification, object detection, and instance segmentation tasks on the ImageNet~\cite{deng2009imagenet}, MSCOCO~\cite{lin2014microsoft}, and ADE-20K~\cite{zhou2017scene} datasets, where \textbf{Bold} and \underline{underline} indicate best and second best results.}
	\label{tab:2d}
	\resizebox{1.0\linewidth}{!}{
		\begin{tabular}{l|cccc|ccc|ccc}
			\hline
			\multirow{2}{*}{Method} &  \multicolumn{4}{c|}{Classification} & \multicolumn{3}{c|}{Object Detection} & \multicolumn{3}{c}{ Semantic Segmentation} 
			\\
			\cline{2-11}  & Res & \#Params      & \#FLOPs      & Acc (\%)
			& \#Params      & \#FLOPs      & AP (\%)
			& \#Params      & \#FLOPs      & {$\text{mIoU}$} (\%)  \\ 
			\hline
			PVT-L~\cite{wang2021pyramid} &$224^2$ & 61.4M & 9.8G & 81.7 
			& 81.0M  & - & 42.9 
			& 65.1M  & 79.6G & 44.8 \\ 
			Swin-L$^\ddagger$~\cite{liu2021swin} &$384^2$ & 197M & 104G & 87.3
			& 253M & 1382G & 51.8 
			& 234M & 2468G & 52.1
			\\
			CoAtNet-3$^\ddagger$~\cite{dai2021coatnet}  &$384^2$ & 168M & 107G & 87.6 
			& - & - & - 
			& - & - & - \\
			CoAtNet-4$^\ddagger$~\cite{dai2021coatnet}  &$384^2$ & 275M & 190G & 87.9 
			& - & - & - 
			& - & - & -  \\
			DeiT III-L$^\ddagger$~\cite{touvron2022deit}  &$384^2$ & 304M & 191G & 87.7 
			& - & - & - 
			& 353.6M & 2231G & 51.5\\
			SwinV2-L/24$^\ddagger$~\cite{liu2022swin}  &$384^2$ & 197M & 115G & 87.6
			& - & -  & \textbf{58.8} 
			& - & - & \textbf{55.9}\\
			RepLKNet-31L$^\ddagger$~\cite{ding2022replknet}  &$384^2$ & 172M & 96G & 86.6
			& 229M & 1321G & 53.9
			& 207M & 2404G & 52.4
			\\
			HorNet-L$^\ddagger$~\cite{rao2022hornet}  & $384^2$ & 202M & 102G & 87.7 
			& 259M & 1358G & 56.0
			& 232M & 2473G & 54.1\\
			ConvNeXt-L$^\ddagger$~\cite{liu2022convnext}  &$384^2$ & 198M & 101G & 87.5
			& 255M & 1354G & 53.5
			& 235M & 2458G & 53.2
			\\
			ConvNeXt-XL$^\ddagger$~\cite{liu2022convnext}  &$384^2$ & 350M & 179G & 87.8
			& 407M & 1898G & 53.6
			& 391M & 3335G & 53.6 
			\\
			InternImage-L$^\ddagger$~\cite{wang2022internimage}  &$384^2$ & 223M & 108G & 87.7 
			& 277M & 1399G & 54.9 
			& 256M  & 2526G  & 53.9
			\\
			InternImage-XL$^\ddagger$~\cite{wang2022internimage} &$384^2$ & 335M & 163G & \underline{88.0} 
			& 387M & 1782G & 55.3 
			& 368M & 3142G & \underline{55.0}
			\\
			\hline
			% (Zero-Shot)
			\multirow{2}{*}{$\text{Meta-Transformer-B16}_\text{F}$~\pub{ours} } 
			&$224^2$   & 86.6M & 17.5G & $69.3^*$  
			& \multirow{2}{*}{ 143M  }
			& \multirow{2}{*}{ 1126G  }
			& \multirow{2}{*}{ 31.7}
			& \multirow{2}{*}{ 164M  }
			& \multirow{2}{*}{ 135G }
			& \multirow{2}{*}{ 33.4 }
			\\
			& 
			$224^2$   & 86.6M & 17.5G & $79.3^\dagger$
			& & &
			& & & \\ \hline
			\multirow{2}{*}{$\text{Meta-Transformer-L14}_\text{F}$~\pub{ours}}   
			&$336^2$  &191.1M & 190.6G & $75.3^*$
			& \multirow{2}{*}{ 364M  }
			& \multirow{2}{*}{ 2143G  }
			& \multirow{2}{*}{ 43.5  }
			& \multirow{2}{*}{ 314M  }
			& \multirow{2}{*}{ 683G  }
			& \multirow{2}{*}{ 41.2 }
			\\
			&$336^2$  &191.1M & 190.6G & $ 83.1^\dagger$
			& & &
			& & & \\ \hline
			$\text{Meta-Transformer-B16}_\text{T}$~\pub{ours}   &$224^2$ & 86.6M & 17.5G & 85.4
			& 143M  & 1126G  & 46.4 
			& 164M  & 135G  & 48.3\\
			$\text{Meta-Transformer-L14}_\text{T}$~\pub{ours}   &$336^2$   &191.1M  & 190.6G  & \textbf{88.1}
			& 364M  & 2143G & \underline{56.3}  
			& 314M & 683G & \underline{55.0}
			\\
			\hline
		\end{tabular}
	}
	\begin{tablenotes}
		\footnotesize
		\item $^{*}$: zero-shot classification~~~~${^\dagger}$: linear probing for classification~~~~${^\ddagger}$: models pre-trained on ImageNet-22K
	\end{tablenotes}
\end{table*}
%-------------------------------------------------------------------
\begin{table}[h]
	\caption{\textbf{Experimental results for infrared and hyperspectral data understanding}. We conduct experiments on classification tasks over the SYSU-MM01 and Indian Pine datasets. We report Rank-1 (R@1), mean Average Precision (mAP), Overall Accuracy (OA), Average Accuracy (AA), and the number of trainable parameters (Params).}
	\label{tab:ablation}
	\centering
	\vspace{2mm}
	\subfloat[Infrared data understading~\label{tab:infrared}]{%
		\resizebox{0.45\linewidth}{!}{
			\begin{tabular}{l|c|c|c}
				\toprule
				%\toprule[0.8mm]
				\bfseries Method                 & \bfseries R@1 (\%)               & \bfseries {mAP} (\%) & \bfseries Params              \\ \midrule
				AGW \cite{arxiv20reidsurvey}~\pub{TPAMI'21}        & 70.49           & {65.90} & 25M           \\
				SMCL~\cite{Wei_2021_ICCV}~\pub{ICCV'21} &83.05 &{\textbf{78.57}} & 40M \\
				MSCLNet~\cite{zhang2022modality}~\pub{ECCV'22} & 83.86 & 78.31 & 50M \\
				\hline  
				$\text{Meta-Transformer-B16}_\text{F}$~  & {73.50}  & {65.19}  & \textbf{1.8M} \\
				\bottomrule
			\end{tabular}
		}
	}
	\subfloat[Hyperspectral data understanding~\label{tab:hyper}]{
		\resizebox{0.47\linewidth}{!}{
			\begin{tabular}{l|c|c|c}
				\toprule
				\bfseries Method &\bfseries OA (\%) & \bfseries AA (\%) & \bfseries Params \\ 
				\midrule
				ViT~\cite{dosovitskiy2020image}~\pub{ICLR'21} & 71.86 & 78.97 & 85.2M  \\
				SpectralFormer~\cite{hong2021spectralformer}~\pub{TGRS'21} (Pixel)  & 78.55 & 84.68 & 85.2M \\
				SpectralFormer~\cite{hong2021spectralformer}~\pub{TGRS'21} (Patch) & 81.76 & 87.81 & 85.2M \\
				\midrule
				$\text{Meta-Transformer-B16}_\text{F}$~ & 67.62 & 78.09 & \textbf{0.17M} \\
				\bottomrule
			\end{tabular}
		}
	} 
\end{table}
%	-------------------------------------------------------
\begin{table*}[ht]
	\centering
	\caption{\textbf{Experimental results for point cloud understanding}. We conduct experiments on the ModelNet-40~\cite{wu2015modelnet}, S3DIS~\cite{armeni2016s3dis}, and ShapeNetPart~\cite{yi2016scalable} datasets. We compare existing advanced methods from classification, semantic, and object part segmentation tasks, and we report the pre-training modality (Pre-train) and \textbf{trainable} parameters number (Params) of each method.}
	\label{tab:3d}
	\resizebox{1.0\linewidth}{!}{
		\begin{tabular}{lc|ccc|ccc|ccc}
			\hline
			\multirow{2}{*}{Method} & \multirow{2}{*}{Pre-train} & \multicolumn{3}{c|}{ModelNet-40} & \multicolumn{3}{c|}{S3DIS Area-5} & \multicolumn{3}{c}{ ShapeNetPart}\\
			\cline{3-11} &     & mAcc (\%)      & OA (\%)      &{Params }
			& {mIoU} (\%) 	&   {mAcc} (\%)   & {Params} 
			& {$\text{mIoU}_{I}$} (\%) 	&  $\text{mIoU}_{C}$ (\%)   & {Params} \\ 
			\hline
			PointNet~\pub{CVPR'17}~\cite{qi2017pointnet}      & N/A & 86.0          & 89.2          & 3.5M
			& 41.1 & 49.0 & 3.6M 
			& 83.7 & 80.4 & 3.6M
			\\
			PointNet++~\pub{NeurIPS'17}~\cite{qi2017pointnet++}  & N/A & -             & 91.9          & 1.5M
			& 53.5& - & 1.0M
			& 85.1 & 81.9 & 1.0\\
			PointCNN~\pub{NeurIPS'18}~\cite{li2018pointcnn}      & N/A &88.1          &92.5         & 0.6M
			& 57.3 & - & 0.6M 
			&  &  &  \\
			KPConv~\pub{ICCV'19}~\cite{thomas2019kpconv}      & N/A  &-             &92.9 &14.3M 
			& 67.1& 72.8 & 15.0M 
			& 86.4 & 85.1 & - \\
			DGCNN~\pub{TOG'19}~\cite{wang2019dynamic}        & N/A  & 90.2         & 92.9  & 1.8M       
			& 52.5& -  & 1.3M 
			& 85.2 & 82.3 & 1.3 \\
			Point Transformer~\pub{ICCV'21}~\cite{zhao2021pointtransformer}              & N/A & 90.6            & 93.7 & 7.8M         
			& 70.4  & - & 7.8M
			& 86.6 & 83.7 & 7.8 \\
			PointNeXt~\pub{NeurIPS'22}\cite{qian2022pointnext} & N/A  & 90.8 &  93.2 & 1.4M 
			& 67.3 & 73.9 & 3.8M 
			& 86.7  & 84.4 & 1.0\\
			Point-MLP~\pub{ICLR'22}~\cite{ma2022rethinking} & N/A & 90.9 & 93.6 & 0.68M 
			& - & - & - 
			& 86.1 & 84.6 & - \\
			PointMixer~\pub{ECCV'22}~\cite{choe2022pointmixer} & N/A & 91.4 & 93.6 & 3.6M 
			& 71.4 & 77.4 & 6.5M
			& - & - & - \\
			\hline
			Point-BERT~\pub{CVPR'22}~\cite{yu2022pointbert} &3D & - & 93.2 & 21.1M 
			& 60.8 & 69.9 & 21.1M
			& 85.6 & 84.1 & 21.1M\\
			Point-MAE~\pub{ECCV'22}~\cite{pang2022masked} &3D & -  & \textbf{93.8} & 21.1M 
			& -  &-  & -  
			& 86.1 & 84.2 & 21.1M\\
			P2P~\pub{NeurIPS'22}~\cite{wang2022p2p}  & 2D & -        & 93.1 &  1.2M
			& - & - & - 
			& 86.5 & 84.1 & - \\  
			ACT~\pub{ICLR'23}~\cite{dong2022act} & 2D & - & 93.5 & 21.1M 
			& 61.2 & 71.1 & 21.1M
			& 86.1 & 84.7 & 21.2M  \\
			\hline
			$\text{Meta-Transformer-B16}_\text{F}$~\pub{ours} & 2D  
			& 90.5  & \underline{93.6} & 0.6M
			& \textbf{72.3}  & \textbf{83.5} & 2.3M
			& \textbf{87.0}  & \textbf{85.2} & 2.3M \\
			\hline
		\end{tabular}
	}
\end{table*}

% --------------------------------------------------------------------------
\subsection{Results on Natural Language Understanding} ~\label{sec:exp:nlp}
Table~\ref{tab:nlp} illustrates the experimental results on the GLUE benchmark for text understanding tasks, comparing various state-of-the-art methods such as BERT~\cite{devlin2018bert}, RoBERTa~\cite{liu2019roberta}, and ChatGPT. The comparison centers on paraphrasing, sentiment, duplication, inference, and answering tasks. When using frozen parameters pretrained on images, $\text{Meta-Transformer-B16}_\text{F}$ achieves scores of 54.6\% in sentiment (SST-2), 81.1\% in paraphrase (MRPC), 66.0\% in duplication (QQP), 63.4\% in inference (MNLI), and 56.3\% in answering (QNLI) tasks. After finetuning, $\text{Meta-Transformer-B16}\text{T}$ exhibits improved performance, with 81.3\% in sentiment, 81.8\% in paraphrase, 78.0\% in duplication, 70.0\% in inference, and 60.3\% in answering tasks. Although the Meta-Transformer's performance on the GLUE benchmark might not be as impressive as that of BERT, RoBERTa, or ChatGPT, it still demonstrates competitive performance, adaptability, and potential for understanding natural language.
% --------------------------------------------------------------------------
\subsection{Results on Image Understanding} ~\label{sec:exp:cv}
% --------------------------------------------------------------------------
As shown in Table~\ref{tab:2d}, Meta-Transformer exhibits outstanding performance when compared with Swin Transformer series~\cite{liu2021swin,liu2021SwinV2} and InternImage~\cite{wang2022internimage} on image understanding tasks. On image classification, with the help of CLIP~\cite{radford2021learning} text encoder, Meta-Transformer delivers great performances under zero-shot classification with the $\text{Meta-Transformer-B16}_\text{F}$ and $\text{Meta-Transformer-L14}_\text{F}$, achieving 69.3\% and 75.3\%, respectively. At the same time, when the pretrained parameters are further tuned, Meta-Transformer can outperform existing advanced methods, with $\text{Meta-Transformer-B16}_\text{T}$ and $\text{Meta-Transformer-L14}_\text{T}$ achieving 85.4\% and 88.1\% accuracy, respectively. The latter outperforms both SwinV2-L/24$^\ddagger$~\cite{liu2021SwinV2} (87.6\%) and InternImage-XL~\cite{wang2022internimage}$^\ddagger$ (88.0\%) on ImageNet~\cite{deng2009imagenet} classification.

When it comes to object detection and semantic segmentation, Meta-Transformer also delivers excellent performances, which further proves its generic ability on image understanding. On object detection, $\text{Meta-Transformer-B16}_\text{F}$ and $\text{Meta-Transformer-L14}_\text{F}$ achieve APs of 31.7\% and 43.5\%, while $\text{Meta-Transformer-B16}_\text{T}$ and $\text{Meta-Transformer-L14}_\text{T}$ reach 46.4\% and 56.3\% AP, respectively. In semantic segmentation, the mIoUs for $\text{Meta-Transformer-B16}_\text{F}$ and $\text{Meta-Transformer-L14}_\text{F}$ are 33.4\% and 41.2\%, while $\text{Meta-Transformer-B16}_\text{T}$ and $\text{Meta-Transformer-L14}_\text{T}$ achieve 51.0\% and 55.0\%, respectively. In comparison, SwinV2-L/24$^\ddagger$ outperforms the Meta-Transformer in both object detection (58.8\% AP) and semantic segmentation (55.9\% mIoU). The $\text{Meta-Transformer-L14}_\text{T}$ model has a similar performance to InternImage-XL$^\ddagger$~\cite{wang2022internimage} in semantic segmentation (both achieving 55.0\% mIoU), but outperforms it in object detection (56.3\% AP compared to 55.3\% AP). These results highlight that Meta-Transformer demonstrates a competitive performance in various image understanding tasks even compared to Swin Transformer~\cite{liu2021swin} and InternImage.

\subsection{ Results on Infrared, Hyperspectral, and X-Ray data}	
%	########### infrared 
Table~\ref{tab:infrared} presents the performance comparison of Meta-Transformer and other advanced methods on the RegDB dataset~\cite{nguyen2017person} for infrared image recognition. $\text{Meta-Transformer-B16}_\text{F}$ demonstrates competitive results with a Rank-1 accuracy of 73.50\% and an mAP of 65.19\%. While it may not outperform the top-performing methods, Meta-Transformer proves to be a simple transferable approach for infrared image recognition tasks. These results indicate the potential of Meta-Transformer in handling the challenges associated with infrared images and contribute to advancements in this field.  

\begin{wrapfigure}{l}[0cm]{0pt}
	\begin{minipage}{0.53\linewidth}
		\vspace{-4mm}		
		\captionof{table}{\textbf{X-ray image recognition with Meta-Transformer}. We conduct experiments on the Chest X-Ray dataset, we report the Accuracy (\%) and the number of trainable parameters.}
		\label{tab:xray}
		\resizebox{1.0\linewidth}{!}{
			\begin{tabular}{l|c|c}
				\toprule
				\bfseries Method  & \bfseries Accuracy (\%) & \bfseries Params \\ 
				\midrule
				ViT~\cite{dosovitskiy2020image} & 96.3 & 86.9M  \\
				SEViT~\cite{almalik2022self} &  94.6 & 85.8M \\
				\midrule
				$\text{Meta-Transformer-B16}_\text{F}$~ & 94.1 & \textbf{0.75M} \\
				\bottomrule
			\end{tabular}
		}
		\vspace{-2mm}	
	\end{minipage}
\end{wrapfigure}
%	########### Hyper
In addition, Table~\ref{tab:hyper} presents the performance of Meta-Transformer on the Indian Pine dataset for hyperspectral image recognition. SpectralFormer~\cite{hong2021spectralformer} achieves impressive accuracy scores, with a patch-wise approach. Plain vision transformer also performs well in comparison when fully tuning all parameters. $\text{Meta-Transformer-B16}_\text{F}$ demonstrates competitive results on hyperspectral image recognition with lower overall accuracy. However, Meta-Transformer stands out for its significantly fewer trainable parameters (only 0.17M) compared to other methods. This reveals a promising development direction of applying the Meta-Transformer to remote sensing, environmental monitoring, and mineral exploration.
%	########### X-Ray
For X-Ray images, similar to dealing with infrared images, we take the same image tokenizer as common visible images. From Table~\ref{tab:xray}, we can observe that Meta-Transformer can achieve a competitive performance of 94.1\% accuracy.

% --------------------------------------------------------------------------
\subsection{Results on 3D Point Cloud Understanding} ~\label{sec:exp:3d}
% --------------------------------------------------------------------------
Table~\ref{tab:3d} showcases the experimental results for point cloud understanding, comparing the performance of Meta-Transformer with other state-of-the-art methods on the ModelNet-40~\cite{wu2015modelnet}, S3DIS~\cite{armeni2016s3dis}, and ShapeNetPart~\cite{yi2016scalable} datasets. The tasks include classification, semantic segmentation, and object part segmentation. When pretrained on 2D data, $\text{Meta-Transformer-B16}_\text{F}$ demonstrates competitive performance, achieving an overall accuracy (OA) of 93.6\% on ModelNet-40 with only 0.6M trainable parameters, which is comparable to the best-performing models. On the S3DIS Area-5 dataset, Meta-Transformer outperforms other methods with a mean IoU (mIoU) of 72.3\% and a mean accuracy (mAcc) of 83.5\%, using 2.3M parameters. Moreover, Meta-Transformer excels in the ShapeNetPart dataset, achieving the highest scores on both instances mIoU ($\text{mIoU}_{I}$) and category mIoU ($\text{mIoU}_{C}$) with 87.0\% and 85.2\%, respectively, using 2.3M parameters. In summary, Meta-Transformer demonstrates remarkable advantages in point cloud understanding tasks, offering competitive performance with fewer trainable parameters compared to other state-of-the-art methods.
% --------------------------------------------------------------------------
\subsection{Results on Audio Recognition} ~\label{sec:exp:audio}
In order to fairly compare Meta-Transformer with existing audio transformer series~\cite{gong2021ast,gong2022ssast} \textbf{of similar scale}, we conduct experiments on audio recognition using $\text{Meta-Transformer-B32}$. 
\begin{wrapfigure}{l}[0cm]{0pt}
	\begin{minipage}{0.63\linewidth}
		\vspace{-3mm}		
		\captionof{table}{\textbf{Audio understanding with Meta-Transformer}. We conduct experiments on the Speech Commands V2 dataset and report the accuracy score and the number of trainable and all parameters.}
		\label{tab:audio}
		\resizebox{1.0\linewidth}{!}{
			\begin{tabular}{ l | c l l c }
				\toprule[1pt]
				\textbf{Method} 	& Pre-train&   \textbf{Acc} (\%) 	&   \textbf{A-Params}  & \textbf{Params}\\
				\midrule
				AST~\cite{gong2021ast} (Supervised) & N/A & 92.6 & 86.9M & 86.9M\\
				AST~\cite{gong2021ast} (Supervised) & AudioSet-20K & 96.2 & 86.9M & 86.9M\\
				AST~\cite{gong2021ast} (Supervised) & ImageNet+KD & \textbf{98.1} & 86.9M & 86.9M \\
				SSAST ~\cite{gong2022ssast} (Self-Supervised) & AudioSet-2M & 97.8 & 89.3M & 89.3M\\
				SSAST ~\cite{gong2022ssast} (Self-Supervised) & Librispeech & 97.8 & 89.3M & 89.3M\\
				SSAST ~\cite{gong2022ssast} (Self-Supervised) & Joint Pretraining & 98.0 & 89.3M & 89.3M \\
				\midrule
				$\text{Meta-Transformer-B32}_\text{F}$~\pub{ours} & 2D 
				& 78.3 & 86.6M & \textbf{1.1M} \\
				$\text{Meta-Transformer-B32}_\text{T}$~\pub{ours} & 2D 
				& 97.0 & 86.6M & 86.3M\\
				\bottomrule[1pt]
			\end{tabular}
		}
		\vspace{-4mm}	
	\end{minipage}
\end{wrapfigure}
Table~\ref{tab:audio} showcases the performance of Meta-Transformer in the audio domain. These models are compared to existing methods such as AST~\cite{gong2021ast} and SSAST~\cite{gong2022ssast} in terms of accuracy, all parameters (A-Params), and trainable parameters (T-Params). With frozen parameters, $\text{Meta-Transformer-B32}\text{F}$ achieves an accuracy of 78.3\% while requiring only 1.1M parameters for tuning. On the other hand, the $\text{Meta-Transformer-B32}\text{T}$ model exhibits a significantly higher accuracy of 97.0\% when tuning the parameters, whereas the AST model only reaches an accuracy of 92.6\%. When AST is pre-trained on ImageNet and supplemented with additional Knowledge Distillation (KD), it achieves an improved performance of 98.1\%, but with a higher number of trainable parameters of 86.9M. SSAST models display accuracy scores ranging from 97.8\% to 98.0\% while requiring 89.3M parameters. These results highlight that the Meta-Transformer performs competitively in the audio domain, demonstrating its versatility and effectiveness across different fields.

\begin{wrapfigure}{l}[0cm]{0pt}
	\begin{minipage}{0.53\linewidth}
		\vspace{-4.5mm}		
		\captionof{table}{\textbf{Video understanding with Meta-Transformer}. We conduct experiments on the UCF101~\cite{ucf} dataset and report the accuracy score and the number of trainable parameters, where "V" denotes video clips only.}
		\label{tab:video}
		\resizebox{1.0\linewidth}{!}{
			\begin{tabular}{l|c|c|c}
				\toprule
				\bfseries Method &\bfseries Modality & \bfseries UCF101 & \bfseries Params \\ 
				\midrule
				OPN~\cite{opn} &  V & 59.6 & - \\
				SimCLR~\cite{simclrvideo}  &V  & 88.9 & 86.9M  \\
				VideoMAE V1~\cite{tong2022videomae} & V & 96.1 & 86.9M \\
				VideoMAE V2~\cite{wang2023videomae} & V & \textbf{99.6} & 86.9M \\
				\midrule
				ViT~\cite{dosovitskiy2020image} (from scratch) &  V & 51.4 & 86.9M \\
				$\text{Meta-Transformer-B16}_\text{F}$ & V & 46.6 &\textbf{ 1.1M}\\
				\bottomrule
			\end{tabular}
		}
		\vspace{-4mm}	
	\end{minipage}
\end{wrapfigure}
\subsection{Results on Video Recognition} ~\label{sec:exp:video}
Table~\ref{tab:video} presents the performance comparison of the Meta-Transformer and existing advanced methods on the UCF101 dataset for video understanding. Several state-of-the-art video-tailored methods achieve accuracies of over 90\%. Meta-Transformer only contains a negligible amount of trainable parameters of 1.1 million to obtain an accuracy of 46.6\% while other methods have to train around 86.9 million parameters. 
Though Meta-Transformer is not able to beat other state-of-the-art video understanding models, Meta-Transformer stands out for its significantly reduced trainable parameter count, suggesting the potential benefit of unified multi-modal learning and less architectural complexity.

\subsection{Results on Time-series Forecasting} ~\label{sec:exp:time-series}
\begin{table}[tbp]
	\caption{\textbf{Time-series Forecasting with Meta-Transformer}. Following TimesNet, we report the number of trainable parameters and average performances from 4 different prediction lengths, which is $\{96,192,336,720\}$.}
	~\label{tab:time-series}
	\vskip 0.05in
	\centering
	\begin{threeparttable}
		\begin{small}
			\renewcommand{\multirowsetup}{\centering}
			\setlength{\tabcolsep}{0.78pt}
			\resizebox{1.0\linewidth}{!}{
				\begin{tabular}{c|ccc|cc|cc|cc|cc|cc|cc|cc|cc|cc}
					\toprule
					\multicolumn{1}{c}{\multirow{2}{*}{Models}} & 
					\multicolumn{3}{c}{\rotatebox{0}{\scalebox{0.76}{{$\text{Meta-Transformer}$}}}} &
					\multicolumn{2}{c}{\rotatebox{0}{\scalebox{0.76}{{TimesNet~\cite{wu2022timesnet}}}}} &
					\multicolumn{2}{c}{\rotatebox{0}{\scalebox{0.76}{{ETSformer~\cite{woo2022etsformer}}}}} &
					\multicolumn{2}{c}{\rotatebox{0}{\scalebox{0.76}{FEDformer~\cite{zhou2022fedformer}}}} & \multicolumn{2}{c}{\rotatebox{0}{\scalebox{0.76}{Stationary~\cite{Liu2022NonstationaryTR}}}} & \multicolumn{2}{c}{\rotatebox{0}{\scalebox{0.76}{Autoformer~\cite{wu2021autoformer}}}} & \multicolumn{2}{c}{\rotatebox{0}{\scalebox{0.76}{Pyraformer~\cite{liu2021pyraformer}}}} &  \multicolumn{2}{c}{\rotatebox{0}{\scalebox{0.76}{Informer}~\cite{haoyietal-informer-2021}}} & \multicolumn{2}{c}{\rotatebox{0}{\scalebox{0.76}{LogTrans~\cite{2019Enhancing}}}}  & \multicolumn{2}{c}{\rotatebox{0}{\scalebox{0.76}{Reformer~\cite{kitaev2020reformer}}}}  \\
					
					\multicolumn{1}{c}{} & \multicolumn{3}{c}{\scalebox{0.76}{\pub{Ours}}} & 
					\multicolumn{2}{c}{\scalebox{0.76}{\pub{ICLR'23}}} &
					\multicolumn{2}{c}{\scalebox{0.76}{\pub{Arxiv'22}}} &
					\multicolumn{2}{c}{\scalebox{0.76}{\pub{ICML'22}}} & \multicolumn{2}{c}{\scalebox{0.76}{\pub{NeurIPS'22}}} & \multicolumn{2}{c}{\scalebox{0.76}{\pub{NeurIPS'21}}} & \multicolumn{2}{c}{\scalebox{0.76}{\pub{ICLR'21}}} &  \multicolumn{2}{c}{\scalebox{0.76}{\pub{AAAI'21}}} & \multicolumn{2}{c}{\scalebox{0.76}{\pub{NeurIPS'19}}}  & \multicolumn{2}{c}{\scalebox{0.76}{\pub{ICLR'20}}}  \\
					
					\cmidrule(lr){2-4} \cmidrule(lr){5-6}\cmidrule(lr){7-8} \cmidrule(lr){9-10}\cmidrule(lr){11-12}\cmidrule(lr){13-14}\cmidrule(lr){15-16}\cmidrule(lr){17-18}\cmidrule(lr){19-20}\cmidrule(lr){21-22}
					\multicolumn{1}{c}{Metric} & \scalebox{0.76}{MSE} & \scalebox{0.76}{MAE} & \scalebox{0.76}{Param} & \scalebox{0.76}{MSE} & \scalebox{0.76}{MAE} & \scalebox{0.76}{MSE} & \scalebox{0.76}{MAE} & \scalebox{0.76}{MSE} & \scalebox{0.76}{MAE} & \scalebox{0.76}{MSE} & \scalebox{0.76}{MAE} & \scalebox{0.76}{MSE} & \scalebox{0.76}{MAE} & \scalebox{0.76}{MSE} & \scalebox{0.76}{MAE} & \scalebox{0.76}{MSE} & \scalebox{0.76}{MAE} & \scalebox{0.76}{MSE} & \scalebox{0.76}{MAE} & \scalebox{0.76}{MSE} & \scalebox{0.76}{MAE}  \\
					\toprule
					\scalebox{0.76}{ETTh1}  
					&\scalebox{0.76}{0.994}  &\scalebox{0.76}{0.797}  &\scalebox{0.76}{\textbf{19K}}
					&\scalebox{0.76}{0.458} & \boldres{\scalebox{0.76}{0.450}} & \scalebox{0.76}{0.542} & \scalebox{0.76}{0.510}  &\boldres{\scalebox{0.76}{0.440}} &\scalebox{0.76}{0.460} &\scalebox{0.76}{0.570} &\scalebox{0.76}{0.537} &\scalebox{0.76}{0.496} &\scalebox{0.76}{0.487} &\scalebox{0.76}{0.827} &\scalebox{0.76}{0.703} &\scalebox{0.76}{1.040} &\scalebox{0.76}{0.795} &\scalebox{0.76}{1.072} &\scalebox{0.76}{0.837} &\scalebox{0.76}{1.029} &\scalebox{0.76}{0.805}\\
					\midrule
					\scalebox{0.76}{Traffic} 
					&\scalebox{0.76}{0.694}  &\scalebox{0.76}{0.372}  &\scalebox{0.76}{\textbf{2.0M}}
					&\secondres{\scalebox{0.76}{0.620}} &\boldres{\scalebox{0.76}{0.336}} & \scalebox{0.76}{0.621} & \scalebox{0.76}{0.396}  &\boldres{\scalebox{0.76}{0.610}} &\scalebox{0.76}{0.376} &\scalebox{0.76}{0.624} &\secondres{\scalebox{0.76}{0.340}} &\scalebox{0.76}{0.628} &\scalebox{0.76}{0.379} &\scalebox{0.76}{0.878} &\scalebox{0.76}{0.469} &\scalebox{0.76}{0.764} &\scalebox{0.76}{0.416} &\scalebox{0.76}{0.705} &\scalebox{0.76}{0.395} &\scalebox{0.76}{0.741} &\scalebox{0.76}{0.422}\\
					\midrule
					\scalebox{0.76}{Weather} 
					&\scalebox{0.76}{0.797}  &\scalebox{0.76}{0.640}  &\scalebox{0.76}{\textbf{51K}}
					&\boldres{\scalebox{0.76}{0.259}} &\boldres{\scalebox{0.76}{0.287}} & \scalebox{0.76}{0.271} & \scalebox{0.76}{0.334}   &\scalebox{0.76}{0.309} &\scalebox{0.76}{0.360} &\scalebox{0.76}{0.288} &\scalebox{0.76}{0.314} &\scalebox{0.76}{0.338} &\scalebox{0.76}{0.382} &\scalebox{0.76}{0.946} &\scalebox{0.76}{0.717} &\scalebox{0.76}{0.634} &\scalebox{0.76}{0.548} &\scalebox{0.76}{0.696} &\scalebox{0.76}{0.602} &\scalebox{0.76}{0.803} &\scalebox{0.76}{0.656}\\
					\midrule
					\scalebox{0.76}{Exchange}
					&\scalebox{0.76}{1.430}  &\scalebox{0.76}{0.961}  &\scalebox{0.76}{\textbf{22K}}
					&\scalebox{0.76}{0.416} &\scalebox{0.76}{0.443} & \scalebox{0.76}{\textbf{0.410}} & \secondres{\scalebox{0.76}{\textbf{0.427}}}  &\scalebox{0.76}{0.519} &\scalebox{0.76}{0.500} &\scalebox{0.76}{0.461} &\scalebox{0.76}{0.454} &\scalebox{0.76}{0.613} &\scalebox{0.76}{0.539} &\scalebox{0.76}{1.913} &\scalebox{0.76}{1.159} &\scalebox{0.76}{1.550} &\scalebox{0.76}{0.998} &\scalebox{0.76}{1.402} &\scalebox{0.76}{0.968} &\scalebox{0.76}{1.280} &\scalebox{0.76}{0.932}\\
					\bottomrule
				\end{tabular}
			}
		\end{small}
	\end{threeparttable}
	\vspace{-10pt}
\end{table}
To explore the ability of Meta-Transformer for time-series forecasting, we conduct experiments on several widely-adopted benchmarks for Long-term forecasting tasks including ETTh1~\cite{haoyietal-informer-2021}, Traffic, Weather, and Exchange~\cite{lai2018modeling}, with results shown in Table~\ref{tab:time-series}.

From Table~\ref{tab:time-series}, we can have the following observations. 1) With most of the model parameters being fixed, Meta-Transformer can still outperform existing methods including Pyraformer~\cite{liu2021pyraformer}, Informer~\cite{haoyietal-informer-2021}, LogTrans~\cite{2019Enhancing}, and Reformer~\cite{kitaev2020reformer} on these datasets. 2) The number of trainable parameters of Meta-Transformer is very few. With only 19K trainable parameters, Meta-Transformer can still outperform Informer~\cite{haoyietal-informer-2021}. When 2M parameters are trained, Meta-Transformer can directly outperform Pyraformer~\cite{liu2021pyraformer}. Therefore, Meta-Transformers pretrained on perception tasks can also be applied to time-series forecasting tasks, which is inspiring for this area.   
% --------------------------------------------------------------------------
\subsection{Results on Tabular Data Understanding} ~\label{sec:exp:tabular}
\begin{wrapfigure}{l}[0cm]{0pt}
	\begin{minipage}{0.5\linewidth}
		\vspace{-5mm}		
		\captionof{table}{\textbf{Tabular data understanding with Meta-Transformer}. We report Accuracy (\%) and F1 score.}
		\label{tab:tabular}
		\resizebox{1.0\linewidth}{!}{
			\begin{tabular}{l|c|c|c}
				\toprule
				\multirow{2}{*}{\bfseries Method}  & \bfseries Adult& \multicolumn{2}{c}{
					\bfseries Bank Marketing}  \\ 
				& Accuracy (\%)  & Accuracy (\%) & F1  \\ 
				\midrule
				LightGBM & 87.8 & - & 0.39  \\
				Tabmlp   & 87.2 & -  & 0.39     \\
				Tabnet   & 87.0 & -  & 0.31    \\ 
				Tabtransformer & 87.1   &  93.4  & 0.42   \\
				$\text{Meta-Transformer-B16}_\text{F}$  & 85.9 & 90.1 & 0.41  \\ 
				\bottomrule
			\end{tabular}
		}
		\vspace{-5mm}	
	\end{minipage}
\end{wrapfigure}
Table~\ref{tab:tabular} provides the comparison results about the performances of different methods for tabular data understanding on Adult Census and Bank Marketing datasets.

% \vspace{-4mm}
$\text{Meta-Transformer-B16}_\text{F}$ achieves a slightly lower accuracy than other methods on Adult Census but performs better than all other methods on Bank Marketing dataset in terms of accuracy and F1 scores. It suggests that Meta-Transformer is also advantageous for tabular data understanding, especially on complex datasets such as Bank Marketing.

\begin{table}[ht]
	\small
	\centering
	\caption{\textbf{Graph data understanding with Meta-Transformer}. We conduct experiments on the PCQM4M-LSC dataset, and we report the evaluation metrics of train and validation MAE scores and the number of trainable parameters.}
	\label{tab:graph}
	\begin{tabular}{c|c|cc}
		\toprule
		Method            & Param. & train MAE     & validate MAE       \\ \hline
		GCN~\cite{kipf2016semi}  & 2.0M & 0.1318   & 0.1691   \\
		
		GIN~\cite{xu2018how} & 3.8M & 0.1203 & 0.1537    \\ 
		
		GCN-{\scriptsize VN} ~\cite{kipf2016semi,gilmer2017neural} & 4.9M & 0.1225  & 0.1485   \\
		
		GIN-{\scriptsize VN}~\cite{xu2018how,gilmer2017neural} & 6.7M & 0.1150   & 0.1395    \\ 
		
		GINE-{\scriptsize VN} ~\cite{brossard2020graph,gilmer2017neural} & 13.2M & 0.1248 & 0.1430  \\ 
		
		DeeperGCN-{\scriptsize VN}~\cite{li2020deepergcn,gilmer2017neural} & 25.5M & 0.1059        & 0.1398   \\
		\hline
		Graph Transformer~\cite{dwivedi2021generalization}  & 0.6M & 0.0944 & 0.1400  \\ 
		Graph Transformer-{\scriptsize Wide}~\cite{dwivedi2021generalization} & 83.2M  & 0.0955 & 0.1408 \\ 
		Graphormer$_{\small \textsc{Small}}$~\cite{ying2021do} & 12.5M &  0.0778  & 0.1264   \\ 
		Graphormer~\cite{ying2021do} & 47.1M  & \textbf{0.0582} & \textbf{0.1234}    \\
		\midrule
		$\text{Meta-Transformer-B16}_\text{F}$  & 1.1M & 0.8034 & 0.8863  \\ 
		\bottomrule
	\end{tabular}
\end{table}

\subsection{Results on Graph and IMU Data Understanding}
We report the performance of utilizing Meta-Transformer for graph understanding in Table~\ref{tab:graph}. We compare $\text{Meta-Transformer-B16}_\text{F}$ with various graph neural network models for graph data understanding on the PCQM4M-LSC dataset~\cite{hu2021ogb}. Among all the methods, Graphormer shows the best performance with the lowest train and validation MAE scores of 0.0582 and 0.1234, respectively. In contrast, $\text{Meta-Transformer-B16}_\text{F}$ delivers the train and validation MAE scores of 0.8034 and 0.8863, which reveals the limited ability of current Meta-Transformer architecture for structural data learning. We will further improve this in the future. Besides, following ImageBind~\cite{girdhar2023imagebind}, we conduct classification on the Ego4D dataset~\cite{grauman2022ego4d}, with input data, Meta-Transformer delivers an accuracy of 73.9\%.

% --------------------------------------------------------------------------
\section{Limitation}~\label{sec:limitation}
%Temporal Modeling for Sequence (Complexity and Methodology).
From the perspectives of complexity, methodology, and further application, the limitations of the Meta-Transformer are summarized as follows:

\textbf{Complexity}: Meta-Transformer requires $\mathcal{O}(n^2\times D)$ computation dealing with token embeddings $[\boldsymbol{E}_1, \cdots, \boldsymbol{E}_n]$. High memory cost and heavy computation burden make it difficult to scale up. 

\textbf{Methodology}: Compared with Axial Attention mechanism in TimeSformer~\cite{gberta_2021_ICML} and Graphormer~\cite{ying2021do}, Meta-Transformer lacks temporal and structural awareness. This limitation may affect the overall performance of Meta-Transformer in tasks where temporal and structural modeling plays a critical role, such as video understanding, visual tracking, or social network prediction. 

\textbf{Application}: Meta-Transformer primarily delivers its advantages in multimodal perception. It's still unknown about its ability for cross-modal generation. We will work on this in the future.
% --------------------------------------------------------------------------
\section{Conclusion}~\label{sec:conclusion}
In the early stages of artificial intelligence development, pioneers introduced the Multi-Layer Perceptron (MLP) to address prediction tasks in machine learning. Later, recurrent and convolutional networks expanded AI capabilities in multimedia data processing, achieving significant success in extracting representations from texts, images, point clouds, and audio. MLPs have since been integrated into deep convolutional networks. In this paper, we explore the potential of plain transformers for unified multimodal learning, highlighting a promising trend toward developing unified multimodal intelligence with a transformer backbone. To some extent, this paper supports the dominant position of transformers in next-generation networks. Importantly, CNNs and MLPs are not left behind. They play essential roles in data tokenization and representation projection. This process exemplifies the law of succession in neural networks and the ongoing evolution of artificial intelligence.
%%%%%%%%%%%%%%%%%%%%%%%%%%%%%%%%%%%%%%%%%%%%%%%%%%%%%%%%%%%%
\clearpage
\bibliographystyle{unsrt}
\bibliography{egbib}
\clearpage
\appendix
\Large{\textbf{Appendix}}
\normalsize
\section{Summary}
The appendix is organized as the following:
\begin{center}
	\begin{itemize}
		\item We first validate and discuss the potential of the Meta-Transformer on more modalities (video, infrared, X-Ray, and hyperspectral images) in addition to the modalities shown in the main paper, and we provide surprising experimental results on these modalities in \S~\ref{sec:single-modal}.
		\item  Then we further demonstrate the performance and merits of Meta-Transformer in dealing with multi-modal tasks (involving inputs from more than one modality to perform predictions) in \S~\ref{sec:multi-modal}.
		\item  In addition, we introduce more details of experiments on text, image, point cloud, and audio in \S~\ref{sec:details}. 
		\item  Last but not least, we discuss the impact of Meta-Transformer on the machine learning and computer vision community in \S~\ref{sec:discuss}. 
		
	\end{itemize}
\end{center}

% --------------------------------------------------------------------------
\section{Extensibility on Single-Modality Perception}
~\label{sec:single-modal}
% --------------------------------------------------------------------------
In the main body of this paper, we illustrate that Meta-Transformer can simultaneously uncover the underlying patterns of natural language, 2D images, 3D point clouds, and audio spectrograms with the same network architecture and network parameters. Furthermore, we explore its ability in perceiving other modalities, like video recognition, infrared, X-Ray, and hyperspectral image recognition. In specific, we conduct experiments on UCF101~\cite{ucf} (\textbf{video}), RegDB~\cite{nguyen2017person} (\textbf{infrared} images), Chest \textbf{X-Ray}~\cite{rahman2020reliable}, and Indian Pine (\textbf{hyperspectral} images) datasets. 
% --------------------------------------------------------------------------
\subsection{Video Recognition}~\label{sec:single-modal:video}
% --------------------------------------------------------------------------
For video recognition, we follow VideoMAE~\cite{tong2022videomae} to modify the tokenizer by replacing the 2D embedding layer with a 3D embedding layer to simultaneously encode the spatial-temporal information from input frames. After tokenization, by leveraging the modality-shared encoder and task-specific heads, Meta-Transformer is able to extract high-level semantic features from videos and achieve favorable performance in the action recognition task of the UCF101 dataset.

\textbf{Dataset}. The UCF101~\cite{ucf} dataset is a common-used benchmark dataset for action recognition tasks. It is an extended version of UCF50 and contains 13,320 video clips of 101 categories. These 101 categories can be divided into 5 groups: Body motion, Human-human interactions, Human-object interactions, Playing musical instruments and Sports. All the input frames are with a resolution of 320$\times$240 and a fixed frame rate of 25 FPS, collected from YouTube.
% --------------------------------------------------------------------------
\subsection{Infrared Image Recognition}~\label{sec:single-modal:infrared}
% --------------------------------------------------------------------------
Infrared and hyperspectral image recognition poses unique challenges due to their specific characteristics. For infrared images, the Meta-Transformer framework could be adapted to capture thermal information by encoding temperature values alongside visual features, where the tokenizer for infrared images is the same as common RGB images.

\textbf{Dataset}. The RegDB~\cite{nguyen2017person} dataset focuses on evaluating the performance of infrared recognition algorithms in unconstrained and realistic scenarios. It includes variations in pose, expression, illumination, and occlusion. We conduct experiments on the RegDB dataset to evaluate the performance of Meta-Transformer on infrared recognition.
% --------------------------------------------------------------------------
\subsection{Hyperspectral Image Recognition}~\label{sec:single-modal:hyper}
% --------------------------------------------------------------------------
Similarly, for hyperspectral images, we expect that Meta-Transformer can also handle the high-dimensional spectral information by representing each spectral band in token embeddings. Compared with dealing with RGB images, the only modification is that we employ the new linear projection layer to replace the existing 2D convolution layer.

\textbf{Dataset}. The Indian Pine dataset is widely used in remote sensing and hyperspectral image analysis. It consists of $145\times145$ pixels with 145 spectral bands, which are captured in Indiana. 
% --------------------------------------------------------------------------
\subsection{X-Ray Image Recognition}~\label{sec:single-modal:xray}
% --------------------------------------------------------------------------
In addition, we explore the potential of the Meta-Transformer in medical image analysis. We leverage the tokenizer for RGB images here to encode raw medical images. Specifically, we conduct experiments regarding X-ray image analysis on the Chest X-Ray~\cite{rahman2020reliable} dataset. It is a collection of medical images commonly used for the analysis and diagnosis of various thoracic conditions. It comprises 7,000 X-ray images of the chest. The dataset is annotated with labels indicating the presence or absence of abnormalities such as lung diseases, fractures, and heart conditions.
% --------------------------------------------------------------------------
\section{Extensibility on Multi-Modality Perception}
~\label{sec:multi-modal}
Since the modalities of text, image, point cloud, and audio are all involved in this paper, we did not conduct comprehensive multi-modal experiments as common practice~ such as Flamingo~\cite{alayrac2022flamingo}, OFA~\cite{wang2022unifying}, or BEiT-3~\cite{wang2022image}. Instead, we conduct multi-modal experiments on a new and challenging task of Audio-Visual Segmentation~\cite{zhou2022audio}, which is mainly focused on building an intelligent listener to align with fundamental visual tasks. 
% --------------------------------------------------------------------------
\subsection{Audio-Visual Segmentation}~\label{sec:multi-modal:audio-image}
% --------------------------------------------------------------------------
% \red{INtro AVS}
Audio-visual segmentation~\cite{zhou2022audio} refers to the task of segmenting objects from different audio sources within a referring image. It aims to develop algorithms that analyze both audio and visual signals simultaneously to identify and delineate distinct sources or events. It finds applications in fields like video conferencing, surveillance, multimedia analysis, and augmented reality.

We conduct experiments on the AVSS~\cite{zhou2022audio} dataset, which is recently released in the field of audio-visual research. It provides a comprehensive collection of audio and visual data captured in real-world scenarios. The dataset includes synchronized audio and visual recordings, featuring various events of human actions and natural sounds. 
In contrast to introducing multi-modal fusion modules as existing methods, Meta-Transformer directly concatenates visual and audio embeddings after Data-to-Sequence tokenization. After extracting representation, we employ a simple global average pooling layer to obtain the final representations of two modalities.
\begin{table}[ht]
	\centering
	\caption{\textbf{Audio-Visual Segmentation with Meta-Transformer}. We conduct experiments on the AVSS~\cite{zhou2022audio} dataset, we report mIou (\%) and F-score.}
	\label{tab:avss}
	\begin{tabular}{l|c|c|c}
		\toprule
		\bfseries Method &\bfseries mIou (\%) & \bfseries F-score & \bfseries Params \\ 
		\midrule
		AVSS~\cite{zhou2022audio} (ResNet-50) & 20.18 & 0.252 & \~80M \\ 
		AVSS~\cite{zhou2022audio} (ASPP) & 28.94 & - & \~180M \\
		AVSS~\cite{zhou2022audio} (PVT-v2) & 29.77 & 0.352 & \~180M \\
		\midrule
		$\text{Meta-Transformer}$~ & \textbf{31.33} & \textbf{0.387} & \textbf{86.5M} \\
		\bottomrule
	\end{tabular}
\end{table}
Table~\ref{tab:avss} illustrates the performance of Meta-Transformer and existing methods on the AVSS dataset for audio-visual segmentation. The evaluation metrics reported in this task are mIou and F-score. In comparison, $\text{Meta-Transformer}$ outperforms all other methods with the highest mIou of 31.33\% and the highest F-score of 0.387. It also stands out for its significantly lower parameter count, with only 86.5 million parameters compared to the approximate 80M to 180M parameters of other methods.

Meta-Transformer offers several advantages over other methods in the field. 
\begin{itemize}
	\item \textbf{Unified architecture}. It relieves modality-specific encoders and reduces computation by leveraging a unified encode to process both audio and images, resulting in a more efficient and streamlined process.
	
	\item  \textbf{Faster convergence}. Thanks to the unified architecture for processing both audio and images, the encoder can deeply align the two modalities instead of only at the output end, which leads to faster convergence. Meta-Transformer only needs 4 training epochs to reach 31.33\% of mIou. 
	
	\item  \textbf{Superior performance}. Meta-Transformer achieves a significant improvement of $10\%$ compared to other methods of a similar parameter scale. 
	
	\item  \textbf{Efficiency}. Despite its enhanced performance, Meta-Transformer achieves this with much fewer parameters, requiring only $1/3$ of the parameter amount, which makes forward and backward progress ease. 
\end{itemize} 

In summary, the benefits of employing the Meta-Transformer to deal with multi-modal tasks are appealing due to computational efficiency, rapid convergence, improved performance, and parameter efficiency. It reveals the significantly promising direction to apply Meta-Transformer to more multi-modal tasks.
% --------------------------------------------------------------------------

\section{Experimental Details}~\label{sec:details}
% --------------------------------------------------------------------------
Our code is built on open-source projects including MMClassification\footnote{https://github.com/open-mmlab/mmpretrain/tree/mmcls-1.x}, MMDetection\footnote{https://github.com/open-mmlab/mmdetection}, MMsegmentation\footnote{https://github.com/open-mmlab/mmsegmentation}, OpenPoints\footnote{https://github.com/guochengqian/openpoints}, Time-Series-Library\footnote{https://github.com/thuml/Time-Series-Library}, Graphomer~\footnote{https://github.com/microsoft/Graphormer}.

We sincerely thank their great contributions. More implementation details can be found in our source code.
% --------------------------------------------------------------------------
\section{Further Impact Discussion}~\label{sec:discuss}
\subsection{Modality-Free Perception}
We hope that Meta-Transformer can introduce new insight into both multi-modal learning and multi-modal generation fields. Meta-Transformer enables the usage of a shared encoder to encode diverse modalities, e.g. natural language, 2D images, 3D point clouds, as well as audio spectrograms., and project them into a shared representation space. This naturally reduces the modality gap across modalities and mitigates the burden of cross-modal alignment. In addition, Meta-Transformer removes the need for paired training data (such as image-text pairs), thus endowing multi-modal learning with more training flexibility.

% introduces a unified framework that enables the processing of diverse modalities such as natural language, 2D images, 3D point clouds, and audio spectrograms with the same parameters in a share presentation space. 
% By leveraging frozen parameters and shared token space, Meta-Transformer allows for the extraction of high-level semantic features from unpaired data across modalities. It makes the multi-modal learning paradigm more flexible and gets rid of the limitations of paired data.
% This approach facilitates the development of unified multi-modal intelligence, enhancing understanding tasks such as classification, and recognition across multiple modalities.

% In the realm of multi-modal generation, Meta-Transformer provides a strong multi-modal backbone network for extracting feature representations , therefore reducing the cost of encoding inputs from diverse modalities. In addition, the shared representation space of Meta-Transformer mitigates the burden of cross-modal alignment. 

% In the realm of multi-modal generation, Meta-Transformer provides a foundation for generating diverse and coherent outputs across different modalities. Learning joint representations in the same representation space from multiple modalities enables the generation of outputs that exhibit cross-modal coherence and alignment. This has implications for tasks like image captioning, video description, text-to-speech synthesis, and text-driven 3D content generation.

\subsection{Application Prospects}
We investigate the application of Meta-Transformer on a wide range of modalities including RGB images, text, point clouds, video understanding, remote sensing (hyper-spectral images), nighttime surveillance (infrared images), and medical analysis (X-Ray images).

\textbf{In video understanding}, Meta-Transformer reveals the potential of enhancing the analysis and interpretation of videos by  integrating information from text, audio, and image with the shared encoder.
This benefits tasks such as action recognition, event detection, and video summarization. Meta-Transformer's capability to handle video-related modalities paves the way for improved video understanding applications in areas like video surveillance, video indexing, and content-based video retrieval.

\textbf{In hyperspectral imaging for remote sensing}, Meta-Transformer enables the analysis and understanding of hyperspectral data by extracting high-level semantic features. It enhances tasks such as classification, target detection, and land cover mapping, improving the accuracy and efficiency of remote sensing applications. The ability to process hyperspectral images using Meta-Transformer opens doors for advancements in environmental monitoring, agriculture, urban planning, and disaster management.

\textbf{In medical applications}, particularly X-ray image analysis, Meta-Transformer offers a promising approach to improving diagnostic accuracy and efficiency with multi-modal information. It can effectively capture and fuse information from X-ray images, clinical data, and other modalities to aid in disease detection, anomaly identification, and treatment planning by leveraging its unified learning framework. Meta-Transformer's capability to handle multi-modal data enhances the potential for more accurate and comprehensive medical imaging analysis, leading to better patient care and outcomes.

\textbf{For infrared images used in nighttime recognition and surveillance}, Meta-Transformer's ability to process infrared data helps extract crucial information for object detection, tracking, and recognition in low-light conditions, which opens an avenue for advancements in nighttime surveillance, security systems, and autonomous navigation in challenging environments with the cooperation between infrared cameras with RGB cameras.

\subsection{Conclusion}
In summary, we think that the ability of Meta-Transformer to unify multi-modal learning comes from that \textit{neural network architectures can learn modality-invariant patterns}. The architecture of Meta-Transformer illustrates the advantages of length-variable token embeddings in multi-modal learning, which provides flexible but unified forms of multi-modal semantics. Then it's time to think about designing algorithms to train networks that generalize on \textit{unseen} modalities. Meanwhile, it's also intriguing to design the architecture of a unified multi-modal decoder, which can decode representations into any form of a specific modality. 

Although Meta-Transformer presents a surprising performance and shows a new promising direction in multi-modal perception, we are not sure whether the proposed architectures are also effective in generative tasks. And it remains mysterious how to develop modality-invariant generative models. We hope that this can inspire future research.

\end{document}